\documentclass[sigconf]{acmart}

\usepackage{algorithmic}
\usepackage{setspace}

\usepackage{multicol}  
\usepackage{multirow}
\usepackage{tabularx}
\usepackage{booktabs}
\usepackage{subcaption}
\usepackage{amsmath}
\usepackage{caption}
\usepackage[linesnumbered, ruled]{algorithm2e}
\hypersetup{draft}
\usepackage{enumitem}
\usepackage{bbm}
\DeclareMathOperator*{\argmax}{arg\,max}

\newcolumntype{C}{>{\centering\arraybackslash}X}
\newcolumntype{P}[1]{>{\centering\arraybackslash}p{#1}}

\newcommand{{\method}}{SE-GNN}

\AtBeginDocument{%
  \providecommand\BibTeX{{%
    \normalfont B\kern-0.5em{\scshape i\kern-0.25em b}\kern-0.8em\TeX}}}

\copyrightyear{2021}
\acmYear{2021}
\setcopyright{acmcopyright}\acmConference[CIKM '21]{Proceedings of the 30th ACM
International Conference on Information and Knowledge Management}{November
1--5, 2021}{Virtual Event, QLD, Australia}
\acmBooktitle{Proceedings of the 30th ACM International Conference on Information
and Knowledge Management (CIKM '21), November 1--5, 2021, Virtual Event, QLD,
Australia}
\acmPrice{15.00}
\acmDOI{10.1145/3459637.3482306}
\acmISBN{978-1-4503-8446-9/21/11}

\begin{document}
\fancyhead{}
\title{Towards Self-Explainable Graph Neural Network}

\author{Enyan Dai}
\affiliation{The Pennsylvania State University}
\email{emd5759@psu.edu}

\author{Suhang Wang}
\affiliation{The Pennsylvania State University}
\email{szw494@psu.edu}
\begin{abstract}
\textit{Graph Neural Networks (GNNs)}, which generalize the deep neural networks to graph-structured data, have achieved great success in modeling graphs. However, as an extension of deep learning for graphs, GNNs lack explainability, which largely limits their adoption in scenarios that demand the transparency of models. Though many efforts are taken to improve the explainability of deep learning, they mainly focus on i.i.d data, which cannot be directly applied to explain the predictions of GNNs because GNNs utilize both node features and graph topology to make predictions. There are only very few work on the explainability of GNNs and they focus on post-hoc explanations. Since post-hoc explanations are not directly obtained from the GNNs, they can be biased and misrepresent the true explanations. Therefore, in this paper, we study a novel problem of self-explainable GNNs which can simultaneously give predictions and explanations. We propose a new framework which can find $K$-nearest labeled nodes for each unlabeled node to give explainable node classification, where nearest labeled nodes are found by interpretable similarity module in terms of both node similarity and local structure similarity.  Extensive experiments on real-world and synthetic datasets demonstrate the effectiveness of the proposed framework for explainable node classification.
\end{abstract}




\maketitle

\section{Introduction}
Graph neural networks (GNNs)~\cite{bruna2013spectral,kipf2016semi,hamilton2017inductive} have made remarkable achievements in modeling graph structured data from various domains such as social networks~\cite{hamilton2017inductive,dai2021say}, financial system~\cite{wang2019semi}, and recommendation system~\cite{wang2019knowledge}. The success of GNNs relies on the message-passing. More specifically, the node representations in GNNs will aggregate the information from the neighbors. As a result, the learned representations will capture the node attributes and local topology information to facilitate various tasks, especially the semi-supervised node classification.

Despite the success in modeling graph data, predictions of GNNs are not interpretable for human, which limits the adoption of GNNs in various domains such as credit approval in finance. \textit{First}, as an extension of deep learning on graphs, the high non-linearity of GNNs makes the predictions difficult to understand.
\textit{Second}, GNNs utilize both node attributes and graph topology to give predictions, which leads to additional challenges to interpret the predictions. Though various approaches~\cite{shu2019defend,papernot2018deep} have been studied to give interpretable predictions or explain trained neural networks, most of them are designed for i.i.d data such as images and cannot handle the relational information. Thus, they cannot be directly applied for GNNs which highly rely on relational information in graphs.
Some initial efforts~\cite{ying2019gnnexplainer,luo2020parameterized,huang2020graphlime} have been taken to address this problem. For instance, GNNExplainer~\cite{ying2019gnnexplainer} takes a trained GNN and its predictions as inputs and output crucial node attributes and subgraphs to explain the GNN's predictions. However, all the aforementioned GNN explainers only focus on the post-hoc explanations, i.e., learning an explainer to explain the outputs of a trained  GNN. Since the post-hoc explanations are not directly obtained from the GNNs, they can be biased and misrepresent the true explanations of the GNNs. 
Therefore, it is crucial to develop a self-explainable GNN, which can simultaneously give predictions and explanations.

\begin{figure}
    \centering
    \includegraphics[width=0.68\linewidth]{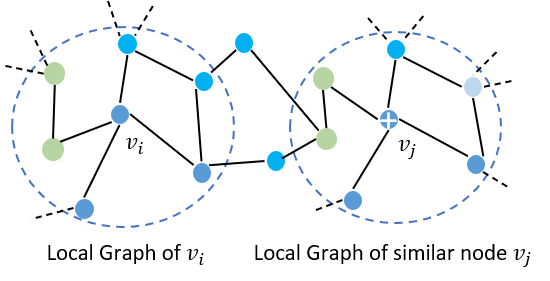}
    \vskip -1.5em
    \caption{Example of interpretable $K$-nearest labeled nodes}
    \label{fig:intro}
    \vskip -1.5em
\end{figure}

One perspective to obtain the self-explanation for node classification is to identify 
\textit{interpretable $K$-nearest labeled nodes} for each node and utilize the $K$-nearest labeled nodes to simultaneously give label prediction and explain why such prediction is given. An example of the interpretable $K$-nearest labeled is shown in Figure~\ref{fig:intro}. As it shows in Figure~\ref{fig:intro}, the topology and node attributes in local graph of node $v_i$ are well matched with that of the node $v_j$ which is labeled as positive. Thus, $v_i$ should be predicted as positive.
Then the explaintions can be: (\textit{i}) ``node $v_i$ is classified as class $X$ because most of the $K$-nearest labeled nodes of $v_i$ belong to class $X$''; and (\textit{ii}) ``node $v_i$ is similar to node $v_j$ because the n-hop subgraphs centered at these two nodes are similar. For example, this part of the structure and attributes of $v_i$'s n-hop subgraph matches to that of $v_j$'s''. Though being promising, the work on exploring $K$-nearest neighbors for self-explainable GNNs is rather limited.

Therefore, in this paper, we investigate a novel problem of self-explainable GNN by exploring $K$-nearest labeled nodes. 
However, developing self-explainable GNN which gives interpretable $K$-nearest labeled nodes is non-trivial. In essense, we are faced with the following challenges: (\textbf{i}) how to incorporate the node similarity and structure similarity to identify $K$-nearest labeled nodes?  For graph-structured data, the similarity should take both node attributes and local topology into consideration. Existing works of interpretable deep KNN~\cite{papernot2018deep} mostly focus on i.i.d data and cannot handle the structure similarity. One straightforward solution is using the cosine similarity of node representations learned by GNNs~\cite{kipf2016semi,hamilton2017inductive} to identify the $K$-nearest labeled nodes. But the selection of the $K$-nearest labeled nodes is not interpretable, because the local graph topology and node attributes are implicitly embedded in latent vectors. In addition, ground-truth of node and structure similarity is generally not available. Due to the lack of explicit supervision, it is challenging to identify the true $K$-nearest labeled nodes; and (\textbf{ii}) how to simultaneously give accurate predictions and correct corresponding explanations. The supervision we have is only the labels for classification accuracy. 
How to leverage label information to help learn the explanations is required to be investigated. 

In an attempt to solve the challenges, we propose a model named as \underline{S}elf-\underline{E}xplainable \underline{GNN} ({\method})\footnote{https://github.com/EnyanDai/SEGNN}. {\method} adopts a novel mechanism that can explicitly evaluate the node similarity and local structure similarity. For the structure similarity estimation, similarity scores of edges across the local graphs of two nodes will be evaluated to provide self-explanations. 
The prediction can be supervised by provided labels. And labels can also provide implicit supervision to guide the similarity modeling, because nodes with same labels are more likely to be a similar pair. 
Therefore, We propose a novel classification loss to simultaneously ensure the accuracy of the prediction and facilitate the interpretable similarity modeling. Furthermore, {\method} adopts contrastive learning on node and edge representations to supervise node similarity and edge matching explanations. The main contributions are:
\begin{itemize}[leftmargin=*]
    \item We study a novel problem of  self-explainable GNN by exploring $K$-nearest labeled nodes for predictions and explanations;
    \item We develop a novel framework {\method}, which adopts an interpretable similarity modeling to identify $K$-nearest labeled nodes and enhance the explanations with self-supervision; 
    \item We generate a synthetic dataset to quantitatively evaluate the quality of explanations, i.e., interpretable $K$-nearest labeled nodes, which can facilitate future research in this direction; and 
    \item We conduct extensive experiments on both real-world datasets and synthetic datasets and demonstrate that {\method} can give accurate predictions and explanations.
\end{itemize}

\section{Related Work}

\subsection{Graph Neural Networks}
Graph Neural Networks (GNNs) have shown their great power in modeling graph-structure data for various applications such as traffic analysis~\cite{zhao2020semi} and drug generation~\cite{bongini2021molecular}. GNNs can be generally split into two categories, i.e., spectral-based~\cite{bruna2013spectral,kipf2016semi,levie2018cayleynets} and spatial-based~\cite{velivckovic2017graph,hamilton2017inductive,chen2018fastgcn,chiang2019cluster,zhao2021graphsmote}. Spectral-based methods learn the node representations based on spectral graph theory. For example, \citeauthor{bruna2013spectral} \cite{bruna2013spectral} first generalize the convolution to graph-structure data with spectral graph theory. To remove the computationally expensive Laplacian eigendecomposition, ChebyNets~\cite{defferrard2016convolutional} define graph convolutions with Chebyshev polynomials. GCN~\cite{kipf2016semi} further simplifies the convolutional operation. Spatial-based graph convolution is defined in spatial domain, which updates node representations by aggregating their neighbors' representations~\cite{niepert2016learning,hamilton2017inductive}. 
For example, GAT \cite{velivckovic2017graph} updates the representations of the nodes from the neighbors with an attention mechanism. Moreover, various spatial methods are proposed for further improvements~\cite{chen2018fastgcn,dai2021nrgnn,chen2020simple,tang2020transferring}. For instance, FastGCN~\cite{chen2018fastgcn} is proposed to solve the scalability issue.

Recently, some works employ self-supervised learning to graph structured data for better node representations. Various pretext tasks have been explored to benefit the training of GNNs~\cite{sun2019multi,zhu2020self,jin2020self}.
For example, SuperGAT~\cite{kim2021find} deploys the edge prediction as self-supervised task to guide the learning of attention. Furthermore, contrastive learning is also widely adopted as the self-supervised task~\cite{sun2019infograph,velivckovic2018deep,qiu2020gcc,you2020graph}. For instance, \citeauthor{qiu2020gcc}~\cite{qiu2020gcc} design a subgraph instance discrimination to better capture the topology information. 
The aforementioned methods are inherently different from our proposed {\method} as we focus on self-explainable GNNs and the self-supervision is applied to obtain high-quality explanations. 

\subsection{Explainability of Graph Neural Networks}
To address the problem of lacking interpretability, extensive works have been proposed. Explanations in the existing works generally fall into two categories, i.e., self-explainable and post-hoc explanations. Self-explainable models are self-intrinsic, which can simultaneously make predictions and explain the predictions~\cite{alvarez2018towards,hind2019ted}. For example, Alvarez et al.~\cite{alvarez2018towards} propose a rich class of interpretable models where the explanations are intrinsic to the model. Post-hoc explanations use another model or strategy to explain the behavior of a trained model, such as  by learning a local approximation with an interpretable model~\cite{ribeiro2016should}, computing saliency map~\cite{zeiler2014visualizing}, identifying feature importance~\cite{selvaraju2017grad, du2018towards,shrikumar2017learning}, obtaining prototypes~\cite{koh2017understanding} and exploring meaning of hidden neurons~\cite{yuan2019interpreting}.

Despite the great success, the aforementioned approaches are overwhelmingly developed for i.id. data such as images and texts; while the work on interpretable GNNs for graph structured data are rather limited~\cite{ying2019gnnexplainer,luo2020parameterized,yuan2020xgnn,huang2020graphlime}. GNNExplainer~\cite{ying2019gnnexplainer} learns soft masks for edges and node features to find the crucial subgraphs and features to explain the predictions. PGExplainer~\cite{luo2020parameterized} applies a parameterized explainer to generate the edge masks from a global view to identify the important subgraphs. GraphLime~\cite{huang2020graphlime} extends the LIME~\cite{ribeiro2016should} to graph neural networks and investigates the importance of different node features for node classification. However, the graph structure information is ignored in GraphLime. 

Our work is inherently different from the aforementioned explainable GNN methods: (i) we focus on learning a self-explainable GNN which can simultaneously give predictions and explanations while all the aforementioned are post-hoc explanations. We don't need additional explainer, which reduces the risk of misrepresenting the true decision reasons of the model; and (ii) we study a novel approach of finding the $K$-nearest labeled nodes in terms of node and structure similarity for classification and explanation.

\section{PROBLEM DEFINITION}
We use $\mathcal{G}=(\mathcal{V},\mathcal{E}, \mathbf{X})$ to denote an attributed graph, where $\mathcal{V}=\{v_1,...,v_N\}$ is the set of $N$ nodes, $\mathcal{E} \subseteq \mathcal{V} \times \mathcal{V}$ is the set of edges, and $\mathbf{X}=\{\mathbf{x}_1,...,\mathbf{x}_N\}$ is the set of node attributes with $\mathbf{x}_i$ being the node attributes of node $v_i$. $\mathbf{A} \in \mathbb{R}^{N \times N}$ is the adjacency matrix of the graph $\mathcal{G}$, where $\mathbf{A}_{ij}=1$ if nodes ${v}_i$ and ${v}_j$ are connected; otherwise $\mathbf{A}_{ij}=0$. In the semi-supervised setting, part of the nodes $\mathcal{V}_L=\{v_1,...,v_l\}\subset\mathcal{V}$ are labeled. We use $\mathcal{Y}=\{\mathbf{y}_1,...,\mathbf{y}_l\}$ to denote the corresponding labels, where $\mathbf{y}_i \in \mathbb{R}^C$ is a one-hot vector of node $v_i$'s label. $\mathcal{V}_U = \mathcal{V}-\mathcal{V}_L$ is a set of unlabeled nodes. 
Most of existing GNNs focus on label prediction of unlabeled nodes in $\mathcal{V}_U$~\cite{wu2020comprehensive}. They usually lack interpretability on why GNN classifiers give such predictions. There are few attempts of post-hoc explainers~\cite{ying2019gnnexplainer,luo2020parameterized,yuan2020xgnn,huang2020graphlime} to provide explanations for trained GNNs; while the work on self-explainable GNNs is rather limited. 

We aim to develop self-explainable GNNs. In particular, for each unlabeled node $v_i$, we want to find $K$ most similar labeled nodes with $v_i$, and simultaneously utilize these similar nodes to predict $v_i$'s label and provide explanation of the prediction. Since both node features and local structure are important for classificaiton, the similarity should be measured in terms of both node features and the n-hop subgraph centered at each node. Sample \textit{explanations can be}: (\textit{i}) ``node $v_i$ is classified as class $B$ because these $K$ nodes have the most similar node features and n-hop subgrph with that of $v_i$, and most of these $K$ nodes belong to class $B$''; (\textit{ii}) ``node $v_i$ is similar to node $v_j$ because the n-hop subgraphs centered at these two nodes are similar. For example, this part of the structure and attributes of $v_i$'s n-hop subgraph matches to that of $v_j$'s''.

Let $\mathcal{G}_s^{(n)}(v_i)=(v_i \cup \mathcal{N}^{(n)}(v_i), \mathcal{E}_s^{(n)}(v_i))$ be the n-hop subgraph centered at node $v_i$, where $\mathcal{N}^{(n)}(v_i)$ is the set of nodes within $n$-hop of $v_i$ and  $\mathcal{E}_s^{(n)}(v_i)$ is the set of edges that linked the nodes in $\mathcal{G}_s^{(n)}(v_i)$. Let $\mathcal{P}^{(n)}(v_i,v_j) \subseteq  \mathcal{E}_s^{(n)}(v_i) \times \mathcal{E}_s^{(n)}(v_j)$ be the edge matching result between the local graphs of node $v_i$ and $v_j$. With these notations, the problem of learning self-explainable GNN by predicting with interpretable $K$-nearest neighbors is defined as:
\vspace*{-0.2em}

\newtheorem{problem}{Problem}
\begin{problem} Given a graph $\mathcal{G}=(\mathcal{V},\mathcal{E})$ with labeled node set $\mathcal{V}_L$ and correspoding label set $\mathcal{Y}$, learn a self-explainable GNN $f_{\mathcal{G}}: \mathcal{G} \rightarrow \mathcal{Y}$ which could give an accurate prediction for each unlabeled node $v_i \in \mathcal{V}_U$ and simultaneously generate explanation from the $K$-nearest labeled node $\{(v_{ik},\mathcal{G}_s^{(n)}(v_{ik}), \mathcal{P}^{(k)}(v_i, v_{ik}))\}_{k=1}^K$, where $v_{ik} \in \mathcal{V}_L$ and $\mathcal{G}_s^{(n)}(v_{ik})$ are the identified top $K$ nearest labeled nodes and the corresponding $n$-hop subgraph, respectively. $\mathcal{P}^{(n)}(v_i, v_{ik})$ is the edge matching results between local graphs of $v_i$ and $v_{ik}$ to explain the similarity in structure. 
\end{problem}

\section{Methodology}
\begin{figure*}
    \centering
    \vskip -1em
    \includegraphics[width=0.95\linewidth]{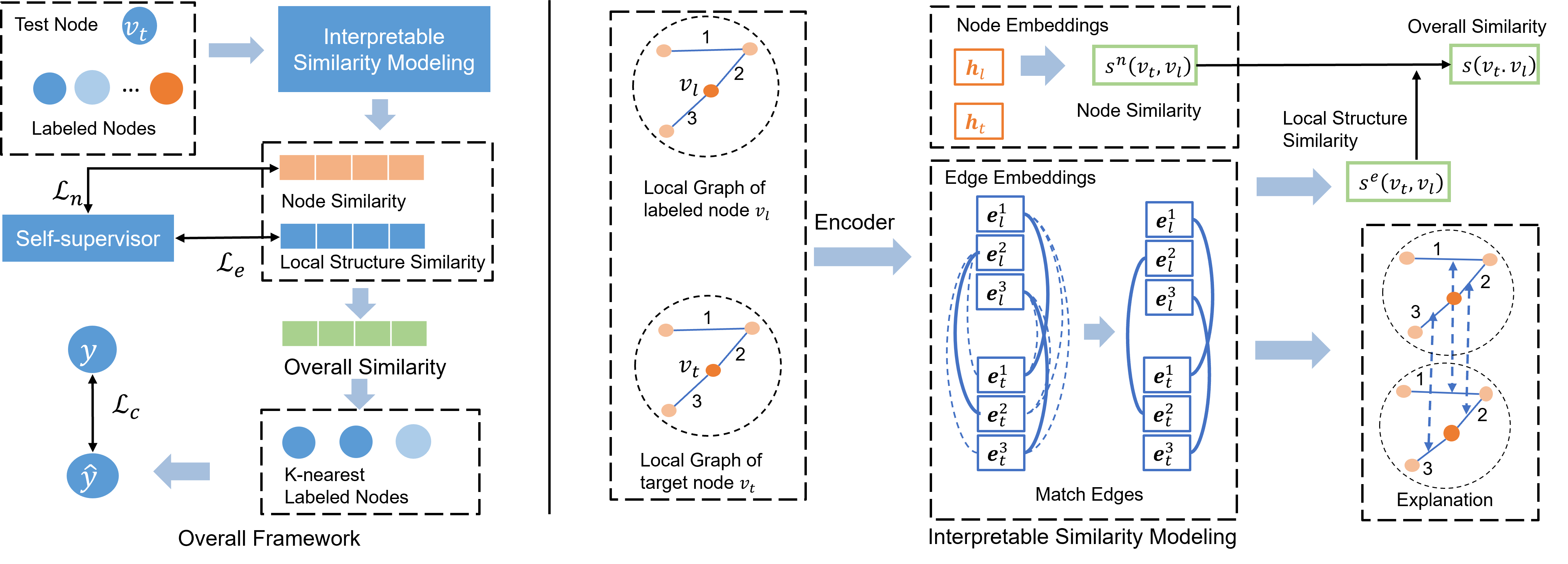}
    \vskip -1.5em
    \caption{An overview of the proposed {\method}.}
    \label{fig:framework}
    \vskip -1em
\end{figure*}

In this section, we present the details of the proposed framework {\method}. The basic idea of {\method} is: for each node $v_i$, it identifies $K$ most similar labeled nodes with $v_i$, and utilize the labels of these $K$ nodes to predict $v_i$'s label. Meanwhile, the $K$ most similar labeled nodes provide explanations on why such prediction is made in terms of both the structure and feature similarity. There are mainly two challenges: (i) how to obtain interpretable $K$-nearest labeled nodes that consider both node and structure similarity; and (ii) how to simultaneously give accurate predictions and correct corresponding explanations. To address these challenges, {\method} explicitly models the node similarity and local structure similarity with explanations. An illustration of the proposed framework is shown in Figure~\ref{fig:framework}. It is mainly composed of an interpretable similarity 
module and a self-supervisor to enhance the explanations. With the novel similarity modeling process, $K$-nearest labeled nodes of the target node and the similarity explanations can be obtained. Then, prediction of the target node can be given based on the identified $K$-nearest labeled nodes.
And, a novel loss function is designed to ensure the accuracy of predictions and facilitate the similarity modeling. Furthermore, self-supervision for explanations is applied to further benefit the accurate explanation generation.
\subsection{Interpretable Similarity Modeling}
\label{sec:4.1}
Since for each node, {\method} relies on interpretable $K$-nearest labeled nodes for predictions and explanations, we need to design an interpretable similarity measurements to measure the similarity of nodes. Unlike i.i.d data, which only needs to measure the similarity from the feature perspective, for graph-structure data, both node attributes and the local graph structures of nodes contain crucial information for node classification. Thus, we propose to explicitly model the node similarity and structure similarity to obtain an interpretable overall similarity. 
\vspace*{-0.2em}

\subsubsection{Node Similarity} The node similarity is to evaluate how similar the target node is with the labeled nodes in the node level. Since node features could be noisy and sparse, directly measuring the node similarity in the raw feature space will result in noisy similarity. Following existing work on similarity metric learning~\cite{qiu2020gcc},  we first learn node representations followed by a similarity function, which could better measure node similarity. 
One straightforward way is to adopt a deep GNN such as GCN~\cite{kipf2016semi} and GAT~\cite{velivckovic2017graph} to learn powerful node embeddings. However, current GNNs are found to experience the over-smoothing issue~\cite{li2018deeper}. This may lead to indistinguishable similarity scores for node pairs that actually differ a lot. And a deep GNN may implicitly model the structure information, which could reduce the interpretability of the node similarity. 
Therefore, we propose to firstly encode the node features with a MLP. Then, the node embeddings are further updated by aggregating the representations from their neighbors with one residual GCN layer~\cite{kipf2016semi}. This process can be mathematically written as:
\begin{equation}
    \mathbf{H}^m = MLP(\mathbf{X}), \quad \mathbf{H} = \sigma(\tilde{\mathbf{A}}\mathbf{H}^{m}\mathbf{W}) + \mathbf{H}^{m},
    \label{eq:node_emb}
\end{equation}
where $\mathbf{X}$ denotes the node attributes, $\tilde{\mathbf{A}}=\mathbf{D}^{-\frac{1}{2}}(\mathbf{A}+\mathbf{I})\mathbf{D}^{-\frac{1}{2}}$ is the normalized adjacency matrix , and $\mathbf{D}$ is a diagonal matrix with $D_{ii}=\sum_{i}A_{ij}$. $\mathbf{I}$ is the identity matrix. $\sigma$ is the activation function such as ReLU. With the learned node embeddings, the node similarity between a target node $v_t$ and a labeled node $v_l$ can be obtained as:
\begin{equation}
    s^n(v_t, v_l) = sim(\textbf{h}_t, \textbf{h}_l),
\end{equation}
where $\textbf{h}_t$ and $\textbf{h}_l$ are the learned embeddings of node $v_t$ and $v_l$, respectively. $sim$ is flexible to be various similarity metrics such as cosine similarity~\cite{you2020graph} and distance-based similarity~\cite{plotz2018neural}.
\vspace*{-0.2em}

\subsubsection{Local Structure Similarity} Generally, the content information, i.e., the n-hop graph structure, is very important for node classification. If two nodes $v_i$ and $v_j$ have similar $n$-hop subgraphs, their labels are likely to be similar. Thus, in addition to the node level similarity, we also measure the local structure similarity, which can (i) explicitly consider the local structure information to facilitate the identification of $K$-nearest labeled nodes and (ii) provide explanations about the similarity in a structure level. Specifically, we propose to measure how well the edges between the local graphs of two nodes match to evaluate the similarity in structure level. The edge matching results can explain the similarity of two nodes in the aspect of local structure. To match edges for local structure similarity, we need to first learn representations of edges. Since an edge is determined by the two nodes linked by it, for an edge $e=(v_i, v_j)$ which links node $v_i$ and $v_j$, we get the edge representation $\mathbf{e}_{ij}$ as:
\begin{equation}
    \mathbf{e}_{ij} = f_e(\mathbf{h}_i, \mathbf{h}_j),
    \label{eq:edge_emd}
\end{equation}
where $\mathbf{h}_i$ and  $\mathbf{h}_j$ are embeddings of node $v_i$ and $v_j$, respectively. $f_e$ is the function of aggregating the node information to get the edge embedding, which is flexible to various functions such as average pooling and LSTM. In our implementation, we apply average pooling function as $f_e$.
With Eq.(\ref{eq:edge_emd}), we can get two sets of edge representations $\mathcal{R}_t=\{\mathbf{e}_t^1,...,\mathbf{e}_t^M\}$ and $\mathcal{R}_l=\{\mathbf{e}_l^1,...,\mathbf{e}_l^N\}$ for $\mathcal{E}_s^{(n)}(v_t)$ and $\mathcal{E}_s^{(n)}(v_l)$, where $\mathcal{E}_s^{(n)}(v_t)$ and $\mathcal{E}_s^{(n)}(v_l)$ are the edges in the local graphs of $v_t$ and $v_l$, respectively.
For an edge $e_t^i \in \mathcal{E}_s^{(n)}(v_t)$, we will find the edge in $\mathcal{E}_s^{(n)}(v_l)$ that matches $e_t^i$ best.
The process of identifying the paired edge for $e_t^i$ can be formally written as:
\begin{equation}
    e_p^i = \argmax_{e_l^j \in \mathcal{E}_s^{(n)}(v_l)}sim(\mathbf{e}_t^i, \mathbf{e}_l^j),
\end{equation}
where $e_p^i$ is the found edge in local graph of $v_l$ that matches edge $e_t^i$.
The whole edge matching results, which can explain the local structure similarity, can be obtained by:
\begin{equation}
    \mathcal{P}^{(n)}(v_t, v_l) = \{(e_t^i, e_p^i)\}_{i=1}^M.
\end{equation}
Then, the local structure similarity between $v_t$ and $v_l$ can be obtained by averaging the similarity scores of all the paired edges:
\begin{equation}
    s^e(v_t, v_l) = \frac{1}{M} {\sum}_{i=1}^M sim(\mathbf{e}_t^i, \mathbf{e}_p^i),
\end{equation}
where $\mathbf{e}_p^i$ is the representation of edge $e_p^i$.
\subsubsection{Overall Similarity} With the node similarity and local structure similarity between node $v_t$ and $v_l$, the overall similarity is:
\begin{equation}
    s(v_t, v_l) = \lambda s^n(v_t, v_l) + (1-\lambda)s^e(v_t, v_l),
    \label{eq:overall_sim}
\end{equation}
where $\lambda$ is a positive scalar to balance the contributions of node similarity and structure similarity.
\subsection{Self-Explainable Classification}
With the similarity metric described in Section~\ref{sec:4.1}, we are able to identify the interpretable $K$-nearest labeled nodes to predict and explain the label of the target node. Next, we introduce the details of the prediction process, discussions about the explanations, and the loss function that ensures classification accuracy.

\subsubsection{Prediction with K-nearest Labeled Nodes} Let $\mathcal{K}_t = \{v_{t}^1,...,v_{t}^K\}$ be the set of $K$-nearest labeled nodes of the target node $v_t$ based on Eq.(\ref{eq:overall_sim}). Intuitively, the more similar $v_t$ and $v_t^i$ are, i.e., the larger $s(v_t, v_t^i)$ is, the more likely $v_t$ has the same label as $v_t^i$. Thus, following deep KNN~\cite{plotz2018neural,ren2014learning},  we predict the label of node $v_t$ as weighted average of the labels of the $K$-nearest neighbors. Specifically, the weight $a_{ti}$  of the $i$-th nearest labeled node, i.e., $v_t^i$, is calculated as
\begin{equation}
    a_{ti} = \frac{\exp(s(v_t,v_t^i)/\tau)}{\sum_{i=1}^{K}\exp(s(v_t, v_t^i)/\tau)},
\end{equation}
where $\tau$ is the temperature parameter.  
With the weight $ a_{ti}$, the label of $v_t$ is predicted as
\begin{equation}
    \hat{\mathbf{y}}_t = {\sum}_{i=1}^K a_{ti}\cdot\mathbf{y}_t^i,
    \label{eq:pred}
\end{equation}
where $\mathbf{y}_t^i$ is the one-hot label vector of $v_t^i$.

\subsubsection{Explanation}
\label{sec:4.2.1_ex}
For a target node $v_t$, the found $K$-nearest labeled nodes $\mathcal{K}_t = \{v_{t}^1,...,v_{t}^K\}$ and the corresponding similarity scores $\{s(v_t, v_{t}^i)\}_{i=1}^K$ can clearly explain the predictive label. 
For the overall similarity $s(v_t, v_t^i)$, the contributions of node and structure similarity can be given through $s^n(v_t, v_t^i)$ and $s^e(v_t, v_t^i)$. Moreover, we can track back the identified edge pairs $\mathcal{P}^{(n)}(v_t, v_t^i)$ between local graph of $v_t$ and its $i$-th nearest labeled node $v_t^i$ to explain the local structure similarity. Though our {\method} focuses on explaining the predictions with $K$-nearest labeled nodes, it also can extract the crucial subgraph for explanation. For a crucial edge, the local graphs of $K$-nearest neighbors should also contain similar one.
Thus, the importance of an edge $e_t^i \in \mathcal{E}_s^{(n)}(v_t)$ can be evaluated by its average similarity with the identified pair edges $\{e_{p}^{ij}\}_{j=1}^K$ in the local graphs of the $K$-nearest labeled nodes as
\begin{equation}
p(e_t^i) = \frac{1}{K} {\sum}_{j=1}^K sim(\mathbf{e}_t^i, \mathbf{e}_p^{ij}),
\end{equation}
where $\mathbf{e}_t^i$ and $\mathbf{e}_p^{ij}$ are edge representations of $e_t^i$ and $e_p^{ij}$, respectively. Then, a threshold can be set to filter out the unimportant edges in the local graph of node $v_t$.

\subsubsection{Classification Loss} {\method} is expected to give accurate predictions. Therefore, we utilize the supervision from the given labels to ensure the accuracy of the self-explainable GNN. One straightforward way is to optimize the predictions of nodes with labels in a leave-one-out manner. More specifically, for a labeled node $v_i \in \mathcal{V}_L$, we identify $K$-nearest labeled nodes from other labeled nodes $\mathcal{V}_L - \{v_i\}$. Then loss such as cross entropy loss can be applied to optimize the model. However, in this way, the optimization only involves the searched $K$-nearest neighbors which are generally similar nodes. Lacking of various negative samples may negatively affect similarity modeling. In addition, the computational cost to identify $K$-nearest neighbors will be large when numerous labeled nodes are given. To address these issues, we propose a loss function that adopts negative sampling~\cite{mikolov2013distributed} and approximately select  $K$-nearest labeled nodes as positive samples. Specifically, for a target node $v_t \in \mathcal{V}_L$, we randomly sample $Q$ labeled nodes $\mathcal{V}_n^t$ which do not share the same label with $v_t$ as negative samples. And we randomly sample $N$ ($N>K$) labeled nodes sharing the same label with $v_t$ as support set to obtain approximate $K$-nearest labeled nodes $\tilde{\mathcal{K}}_t$.
The objective function can be formally written as:
\begin{equation}
    \min_{\theta} \mathcal{L}_c = \frac{1}{|\mathcal{V}_L|}\sum_{v_t \in \mathcal{V}_L}-\log \frac{\sum_{\tilde{v}_t^i \in \tilde{\mathcal{K}}_t} \exp (s(v_t, \tilde{v}_t^i)/\tau)}{\sum_{v_n^i \in \tilde{\mathcal{K}}_t \cup \mathcal{V}_n^t} \exp(s(v_t, v_n^i)/\tau)},
    \label{eq:class}
\end{equation}
where $\theta$ denotes the parameters of {\method}, and $\tau$ is the temperature hyperparameter. With loss function Eq.(\ref{eq:class}), the similarity scores of node pairs with different labels will be minimized, and the similarity scores between a node and its approximate k-nearest neighbor with the same labels will be maximized. As a result, accurate predictions can be given. Moreover, it provides supervision to guide the similarity modeling. \textit{Note that this sampling strategy to obtain approximate $K$-nearest labeled nodes is only applied during the training phase}. For testing phase, $K$-nearest labeled nodes are identified from the whole labeled node set $\mathcal{V}_L$ and the label is predicted with Eq.(\ref{eq:pred}).

\subsection{Enhance Explanation with Self-Supervision}
Although the labels can provide the supervision to facilitate the similarity modeling with objective function in Eq.(\ref{eq:class}), they do not explicitly supervise the node similarity and local structure similarity. In addition, the edge matching results and identification of nearest labeled nodes may not generalize well to unlabeled nodes, because unlabeled nodes are not involved in Eq.(\ref{eq:class}). To further benefit the explanation generation and similarity metric learning, we adopt a contrastive pretext task to provide self-supervision for node similarity and local structural similarity learning. 

Recently, contrastive learning has shown to be effective for unsupervised representation learning on graphs~\cite{velivckovic2018deep,qiu2020gcc,you2020graph}. Essentially, contrastive learning aims to maximize representation consistency under differently augmented views. In other words, with contrastive learning, similar nodes/graphs will be given similar representations. 
Therefore, we adopt contrastive learning on node representations to facilitate the node similarity modeling. Moreover, the explanation for local structure similarity relies on accurate edge machining. To guide the edge matching on unlabeled nodes, a contrastive task on edge representations is deployed as well. In detail, we maximize the the agreement between two augmented views of the graphs via a contrastive loss in node and edge representations. Following GraphCL~\cite{you2020graph}, two augmentations, i.e., attribute masking and edge perturbation, are applied to obtain representations from different views. We apply infoNCE loss~\cite{oord2018representation} as the contrastive loss. The infoNCE loss transfers the mutual information maximization to a classification task which requires positive pairs and negative pairs.
Representations of the same node/edge from these two views will compose positive pairs for contrastive learning. And representations of different nodes in both views compose negative pairs. 
The contrastive learning is trained in a minibatch manner. For a query representation $\mathbf{h}_i$, a dictionary $\{\tilde{\mathbf{h}}_0,..., \tilde{\mathbf{h}}_{Q_N} \}$ which contains one positive sample $\tilde{\mathbf{h}}_+$ will be built. The objective function of contrastive learning on node representations can be formulated as: 
\begin{equation}
    \min_{\theta} \mathcal{L}_n =\frac{1}{|\mathcal{V}_B|} \sum_{v_i \in \mathcal{V}_B} -\log \frac{\exp(sim(\mathbf{h}_i, \tilde{\mathbf{h}}_+)/\tau)}{\sum_{j=0}^{Q_N} \exp(sim(\mathbf{h}_i, \tilde{\mathbf{h}}_j)/\tau)}.
    \label{eq:cont_node}
\end{equation}

Similarly, contrastive learning is adopted on the edge representations to benefit the edge matching explanations. Let $\mathcal{E}_B$ denotes a minibatch of edges. The objective function can be written as:
\begin{equation}
    \min_{\theta} \mathcal{L}_e = \frac{1}{|\mathcal{E}_B|} \sum_{e_i \in \mathcal{E}_B} -\log \frac{\exp(sim(\mathbf{e}_i, \tilde{\mathbf{e}}_+)/\tau)}{\sum_{j=0}^{Q_E} \exp(sim(\mathbf{e}_i, \tilde{\mathbf{e}}_j)/\tau)},
    \label{eq:cont_edge}
\end{equation}
where $\mathbf{e}_i$ denotes a query edge representation, and $\tilde{\mathbf{e}}_+$ is the positive sample from dictionary $\{\tilde{\mathbf{e}}_0,..., \tilde{\mathbf{e}}_{Q_E} \}$.
With the self-supervision, the representations of similar edges and nodes will be similar, which can facilitate the similarity modeling for explanations. Moreover, since edge perturbation is used to augment the graph in contrastive learning, the representations from the perturbed graph will be enforced to be consistent with the representations from the clean graph. This will lead to robustness against structure noise. 

\subsection{Overall Objective Function}
With the supervision from the labels for accurate prediction, and the contrastive learning on node and edge representations to facilitate the modeling of similarity, the final loss function of {\method} is:
\begin{equation}
    \min_{\theta} \mathcal{L}_c + \alpha \mathcal{L}_n +\beta \mathcal{L}_e,
    \label{eq:overall}
\end{equation}
where $\theta$ represents the parameters of {\method}. $\alpha$ and $\beta$ are hyperparameters that control the contributions of self-supervision on node similarity modeling and edge matching in local similarity evaluation, respectively. 

\subsection{Training Algorithm and Time Complexity}

\subsubsection{Training Algorithm} 
The training algorithm of {\method} is given in Algorithm~\ref{alg:1}. In line 1, the parameters of {\method} are randomly initialized with Xavier initialization~\cite{glorot2010understanding}. In
line 3, the supervised loss from the labeled nodes is calculated. Graph augmentation is conducted in line 4, which gives graphs in different views for contrastive learning. From line 5 to line 9, contrastive losses for nodes and edges are computed. For the size of negative samples in Eq.(\ref{eq:class}), it is fixed as 20. The sizes of negative samples in Eq.(\ref{eq:cont_node}) and Eq.(\ref{eq:cont_edge}) are both set as 100 during the training phase.

\begin{algorithm}[t] 
\caption{ Training Algorithm of {\method}.} 
\begin{algorithmic}[1]
\REQUIRE
$\mathcal{G}=(\mathcal{V},\mathcal{E}, \mathbf{X})$ , $\mathcal{Y}$, $K$, $\alpha$, $\beta$, $\lambda$, $\tau$.
\ENSURE Self-explainable GNN $f_{\mathcal{G}}$.
\STATE Randomly initialize the parameters of $f_{\mathcal{G}}$.
\REPEAT
\STATE Obtain the classification loss by Eq.(\ref{eq:class})
\STATE Augment the $\mathcal{G}$ with attribute masking and edge perturbation to receive graphs in different views 
\STATE Sample a node batch $\mathcal{V}_B$ with positive and negative pairs
\STATE Calculate contrastive loss on nodes by Eq.(\ref{eq:cont_node})
\STATE Sample an edge batch $\mathcal{E}_B$ with positive and negative pairs
\STATE Get contrastive loss on edges by Eq.(\ref{eq:cont_edge})
\STATE Optimize the parameter of $f_\mathcal{G}$ by Eq.(\ref{eq:overall})
\UNTIL convergence
\RETURN $f_{\mathcal{G}}$
\end{algorithmic}
\label{alg:1}
\end{algorithm}

\subsubsection{Time Complexity} 
During the test phase, the main time complexity comes from the similarity scores calculation between the test node $v_t$ and all the labeled nodes $\mathcal{V}_L$. For each node $v_l \in \mathcal{V}_L$, the cost for calculating the edge matching with $v_t$ is $\mathcal{O}(d\cdot|\mathcal{E}_t|\cdot|\mathcal{E}_l|)$, where $d$ is the embedding dimension. Thus, the time complexity for one test node is approximately $\mathcal{O}(d\cdot|\mathcal{E}_t| \cdot \sum_{v_l \in \mathcal{V}_L}|\mathcal{E}_l|)$.
As for the training phase, we adopt a sampling strategy in Eq.(\ref{eq:class}) to reduce the pairs of similarity to be computed. For each node $v_t \in \mathcal{V}_L$, let $\mathcal{S}_t = \mathcal{K}_t \cup \mathcal{V}_n^t$ denotes the sampled positive and negative nodes in Eq.(\ref{eq:class}), the cost of calculating classification loss for $v_t$ is $\mathcal{O}(d\cdot|\mathcal{E}_t|\cdot \sum_{v_l \in \mathcal{S}_t} \cdot|\mathcal{E}_l|)$. Thus, the cost for classification loss of all labeled nodes is $\mathcal{O}(\sum_{v_t \in \mathcal{V}_t} \sum_{v_l \in \mathcal{S}_t} d\cdot|\mathcal{E}_t| \cdot|\mathcal{E}_l|)$.
With the computation cost on contrastive learning, the overall time complexity for an iteration in the training phase is $\mathcal{O}(d\cdot (Q_n|\mathcal{V}_B|+ Q_e|\mathcal{E}_B|+\sum_{v_t \in \mathcal{V}_t} \sum_{v_l \in \mathcal{S}_t} |\mathcal{E}_t| \cdot|\mathcal{E}_l|))$. 
\section{experiments}
In this section, we conduct extensive experiments on real-world and synthetic datasets to demonstrate the effectiveness of {\method}.
In particular, we aim to answer the following research questions:
\begin{itemize}[leftmargin=*]
    \item \textbf{RQ1} Can our proposed method simultaneously provide accurate predictions and corresponding reasonable explanations?
    \item \textbf{RQ2} Is {\method} robust to the structure noises in the datasets?
    \item \textbf{RQ3} How does each component of our proposed {\method} contribute to the classification performance and explainability?
\end{itemize}
\subsection{Datasets}
To quantitatively and qualitatively evaluate {\method} in predictions and explanations, we conduct extensive experiments on three real-world datasets and two synthetic datasets. The statistics of the datasets are presented in Table~\ref{tab:dataset}.

\subsubsection{Real-World Datasets} To demonstrate the effectiveness of our proposed methods, for real-world datasets, we choose three widely used benchmark networks, i.e., Cora, Citeseer, and Pubmed~\cite{sen2008collective}. For Cora and Pubmed, the standard dataset splits as in the cited paper are applied. As for Citeseer, it contains isolated nodes which are not applicable to our method. Therefore, following the pre-processing strategy in~\cite{jin2020graph}, we select the largest connected component in the Citeseer graph and apply the same dataset splits.

\subsubsection{Synthetic Datasets} We construct two synthetic datasets, i.e., Syn-Cora and BA-Shapes, which provide ground truth of explanations for quantitative analysis. The details are described below.

\noindent \textbf{Syn-Cora}: This dataset is synthesized from the Cora graph which provides ground-truth of explanations, i.e., $K$-nearest labeled nodes and edge machining results. To construct the graph, motifs are obtained by sampling local graphs of nodes from Cora. Various levels of noises are applied to the motifs in attributes and structures to generate similar local graphs. 
For a motif and it's corresponding perturbed versions provide the groundtruth $K$-nearest neighbors and the corresponding edge-matching.
Specifically, we sample three motifs for each class, resulting in 21 unique motifs in total. To link the synthetic local graphs together, a subgraph of Cora that have no overlap with the motifs is sampled as the basis graph. 
These synthetic local graphs are attached to the basis graph by randomly linking three nodes. To simulate a realistic training scenario, we randomly select 30\% nodes from the motifs and basis graph as the training set. Testing is conducted on the remaining nodes in the motifs for explanation accuracy evaluation.  

\noindent \textbf{BA-Shapes}: To compare with the state-of-the-art GNN explainers~\cite{ying2019gnnexplainer,luo2020parameterized} which identify crucial subgraphs for predictions, we construct BA-Shapes following the setting in GNNExplainer~\cite{ying2019gnnexplainer}. BA-Shapes is a single graph consisting of a base Barabasi-Albert (BA) graph with 300 nodes and
80 “house”-structured motifs. These motifs are attached to the BA graph. And random edges are added to perturb the graph. Node features are not assigned in BA-Shapes. Nodes in the base graph are labeled with 0. Nodes locating at the top/middle/bottom of the “house” are labeled with 1,2,3, respectively. Dataset split is the same as that in~\cite{ying2019gnnexplainer}.

\begin{table}[!t]
    \small
    \caption{Statistics of datasets.}
    \vskip-1.5em
    \centering
    \begin{tabularx}{0.85\linewidth}{p{0.19\linewidth}CCCCC}
    \toprule
         & Nodes & Edges & Features & Classes \\
    \midrule
    Cora & 2,708  & 5,429 & 1,433 & 7\\
    Citeseer & 2,110  & 3,668 & 3,703 & 6\\
    Pubmed & 19,171 & 44,338  & 500 & 3\\
    Syn-Cora & 1,677 & 4,610 & 1,433 & 7 \\
    BA-Shapes & 700 & 4,421 & - & 4  \\
    \bottomrule
    \end{tabularx}
    \label{tab:dataset}
    \vskip -1.5em
\end{table}



\begin{table*}[t]
    \small
    \centering
    \caption{Node classification accuracy (\%) on real-world datasets.}
    \vskip -1.5em
    \begin{tabularx}{0.88\linewidth}{CCCCCCCCC}
    \toprule
    Dataset & GCN & GIN & SuperGAT & Pro-GNN & MLP-K & GCN-K & GIN-K & Ours\\
    \midrule
    Cora & 80.8$\pm 1.1$ & 80.5$\pm 0.8$ & \textbf{82.4}$\pm \mathbf{0.7}$ & 79.1$\pm 0.1$ & 53.9$\pm 1.8$ & 78.8$\pm 1.2$ & 78.8$\pm 0.3$ & 80.4$\pm 0.3$ \\
    Citeseer & 71.9$\pm 1.0$ & 72.5$\pm 0.8$ & 73.6$\pm 0.2$ & 73.3$\pm 0.7$ &61.6$\pm 1.6$ & 71.4$\pm 0.8$ & 69.2$\pm 1.2$ & \textbf{73.8}$\pm \mathbf{0.6}$  \\
    Pubmed & 78.4$\pm 0.4$ & 78.9$\pm 0.2$ & 79.2$\pm 0.4$ & 79.4 $\pm 0.4$ &73.2$\pm 0.2$ & 77.4$\pm 0.2$ & 78.8$\pm 0.3$ & \textbf{80.0}$\pm \mathbf{0.2}$ \\
    \bottomrule
    \end{tabularx}
    
    \label{tab:real_cls}
    \vskip -1.2em
\end{table*}

\subsection{Experimental Settings}
\subsubsection{Baselines} 
To evaluate the performance and robustness of {\method} in node classification on real-world datasets, we first compare with the following representative and state-of-the-art GNNs, self-supervised GNN, and robust GNN.

\begin{itemize}[leftmargin=*]
    \item \textbf{GCN}~\cite{kipf2016semi}: GCN is a popular  spectral-based GNN which defines graph convolution with  spectral theory.
    \item \textbf{GIN}~\cite{xu2018powerful}: Compared with GCN, multi-layer perception is used in GIN to process the aggregated information from the neighbors in each layer to learn more powerful representations.
    \item \textbf{SuperGAT}~\cite{kim2021find}: This is a self-supervised graph neural network. Edge prediction is deployed as the pretext task to directly guide the learning of attention to facilitate the information aggregation.
    \item \textbf{Pro-GNN}~\cite{jin2020graph}: This is state-of-the-art GNN against noisy edges in graphs. It applies low-rank and sparsity constraint to directly learn a clean graph close to the noisy graph to defend against adversarial attacks. 
\end{itemize}
Existing work that identifies the $K$-nearest labeled nodes is rather limited. Therefore, we compare with the following baselines based on the deep KNN~\cite{papernot2018deep} on i.i.d data to evaluate our explanations. 
\begin{itemize}[leftmargin=*]
    \item \textbf{MLP-K}: Following~\cite{papernot2018deep}, we first  train a MLP with the node features and labels. After training, the node representations from the final layer of MLP is used to find the $K$-nearest labeled nodes. The final prediction of an unlabeled node is obtained using weighted average of the its $K$-nearest labeled nodes.
    \item \textbf{GCN-K}: Similar to MLP-K, we use the representations from the final layer of GCN to get $K$-nearest neighbors and predictions.
    Post-hoc explanations in structure similarity can be obtained by edge matching through the edge representations. Edge representations are average representations of the linked nodes .
    \item \textbf{GIN-K}: It replaces the backbone in GCN-K with GIN to obtain the predictions and explanations.
\end{itemize}
Finally, we compare with the state-of-the-art GNN explainers in extracting important subgraphs to explain predictions:
\begin{itemize}[leftmargin=*]
    \item GNNExplainer~\cite{ying2019gnnexplainer}: GNNExplainer takes a trained GNN and the predictions as input to obtain post-hoc explanations. It learns a soft edge mask for each instance to identify the crucial subgraph. 
    \item PGExplainer~\cite{luo2020parameterized}: It adopts a MLP-based explainer to obtain the important subgraphs from a global view to reduce the computation cost and obtain better explanations.
\end{itemize}

\subsubsection{Implementation Details} 
\label{Sec:5.2.2_imp}
\textit{The local structure similarity is based on 2-hop local graphs of nodes in all the experiments.}
For {\method}, the encoder consists of two MLP layers and one GCN layer with residual connection. The hidden dimension is set as 64 in these MLP and GCN layers. 
The rate of attribute masking in the contrastive learning is set as 0.2. We replace 10\% edges in the graph to noisy edges for edge perturbation.
For experiments on real-world datasets, all hyperparameters are tuned based on the prediction results on the validation set. We vary $\alpha$ and $\beta$ among $\{0.0001, 0.001, 0.01, 0.1, 1\}$.
The $\lambda$ which balance the node similarity and structure similarity is set as 0.5 for all datasets. The number of nearest labeled nodes used for prediction, i.e., $K$, is set as searched as $\{25, 40, 50\}$ for all the datasets.  The temperature hyperparameter $\tau$ is fixed as 1 in all experiments.
For the hyperparameters on the Syn-Cora and BA-Shapes, we reuse the setting on the Cora graph. The BA-shapes dataset does not provide features, we initialize the node features with node degree and number of involved triangles. The hyparameters for the baselines are also tuned on the validation set.
All the experiments are conducted 5 times and the average results with standard deviations are reported.

\begin{figure}[t]
\centering
\begin{subfigure}{0.49\columnwidth}
    \centering
    \includegraphics[width=0.9\linewidth]{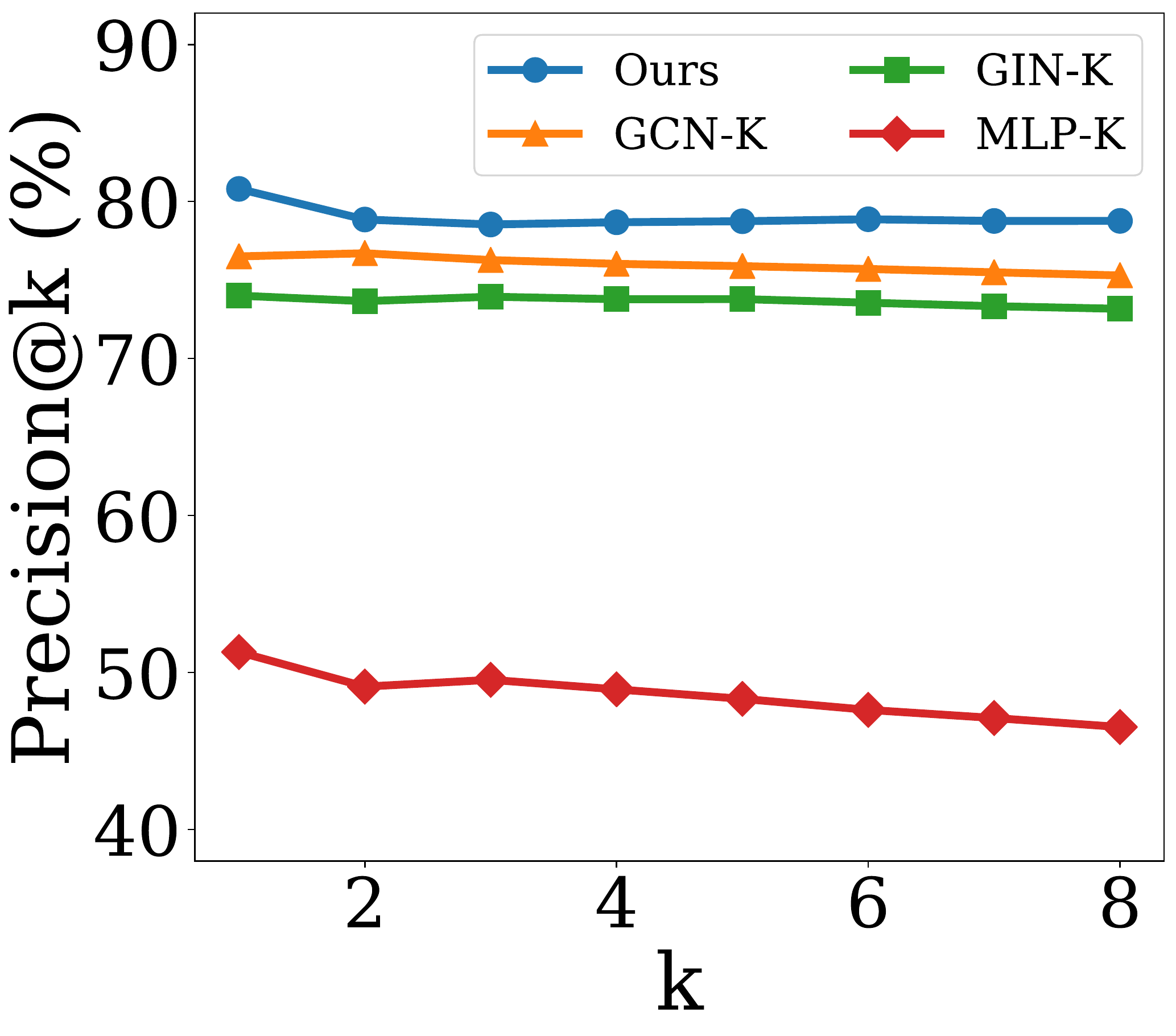} 
    \vskip -0.8em
    \caption{Cora}
    \label{fig:pre_cora}
\end{subfigure}
\begin{subfigure}{0.49\columnwidth}
    \centering
    \includegraphics[width=0.9\linewidth]{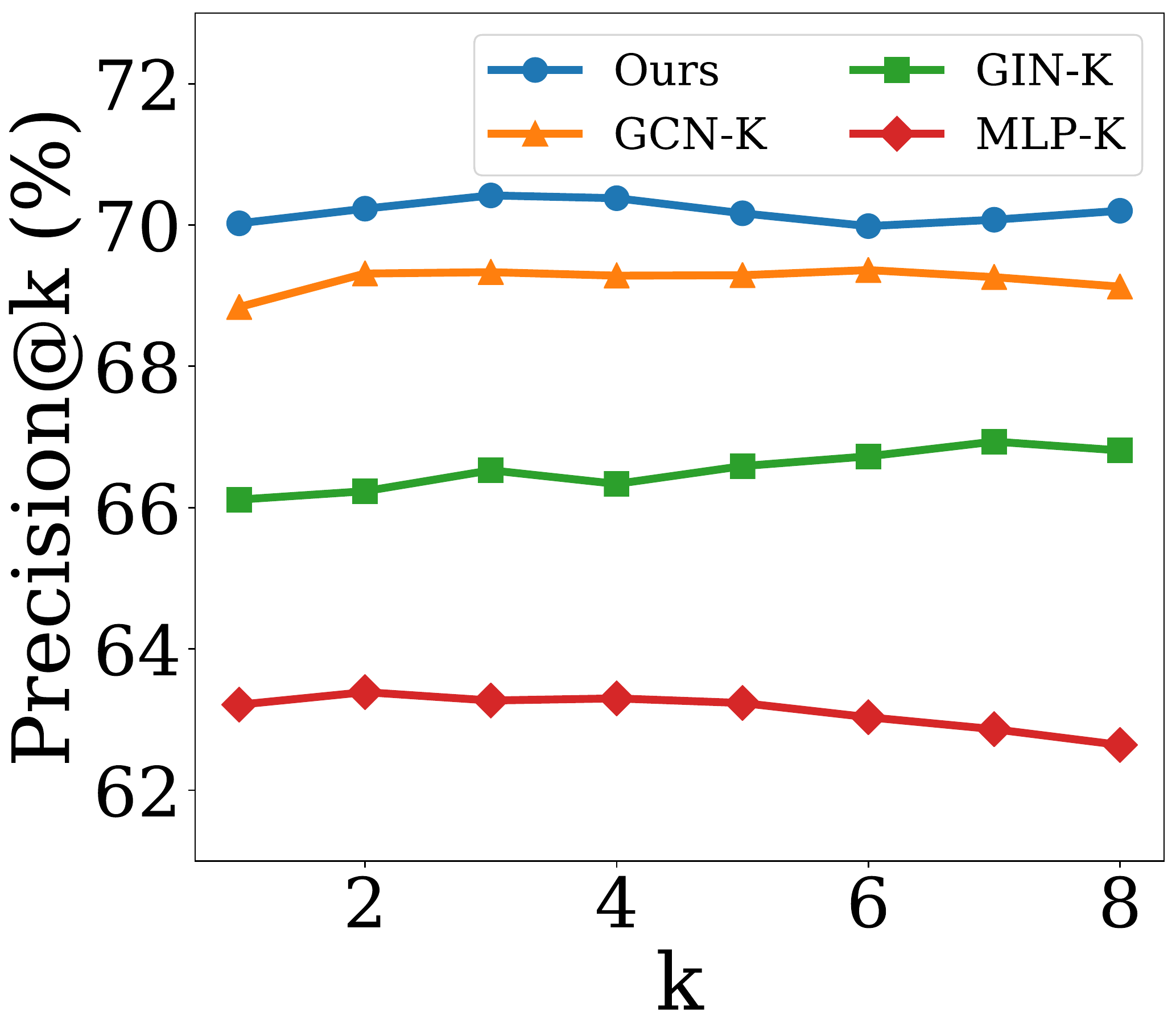} 
    \vskip -0.8em
    \caption{Citeseer}
    \label{fig:pre_citeseer}
\end{subfigure}
\begin{subfigure}{0.49\columnwidth}
    \centering
    \includegraphics[width=0.9\linewidth]{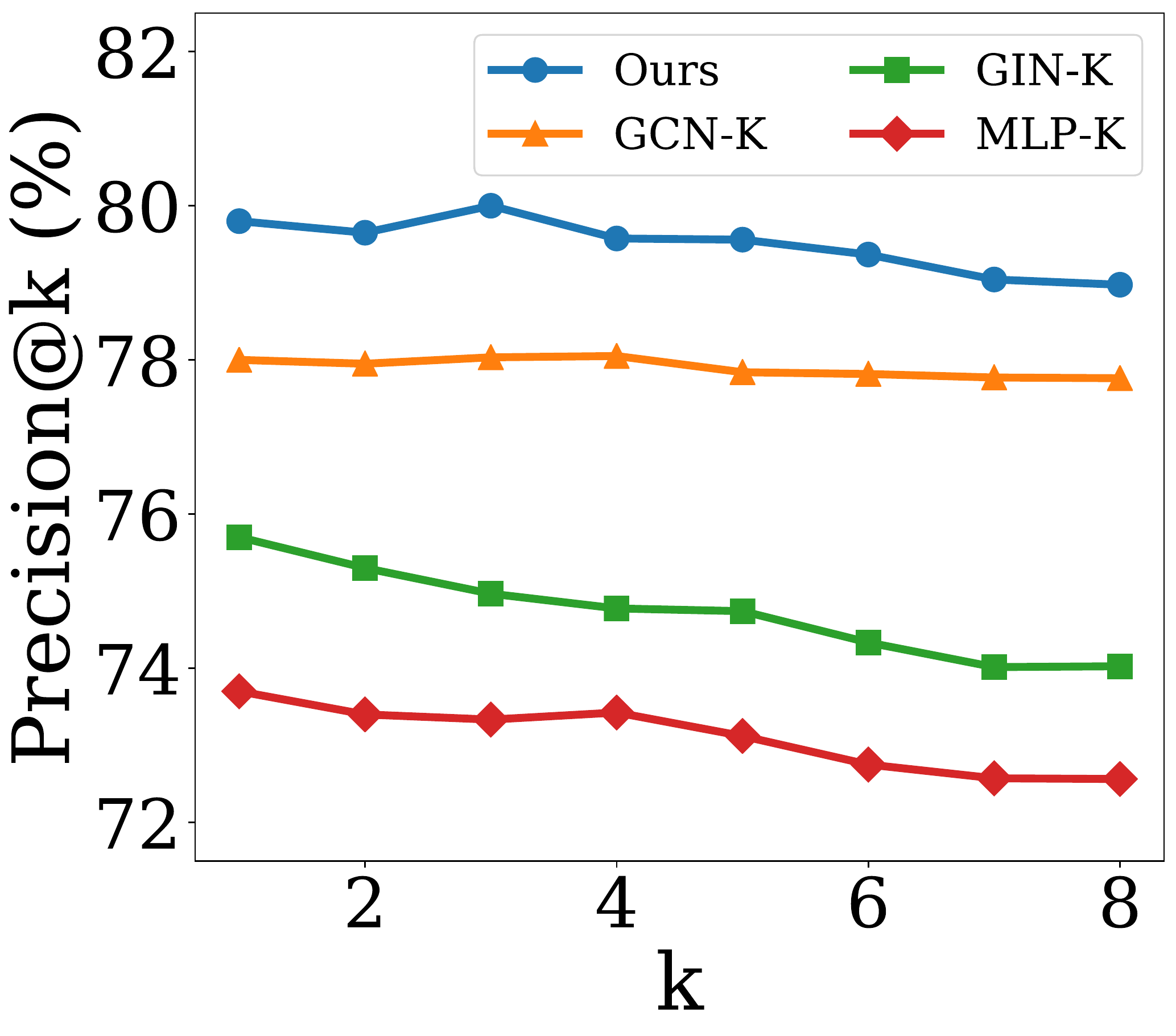} 
    \vskip -0.8em
    \caption{Pubmed}
    \label{fig:pre_pubmed}
\end{subfigure}
\begin{subfigure}{0.49\columnwidth}
    \centering
    \includegraphics[width=0.9\linewidth]{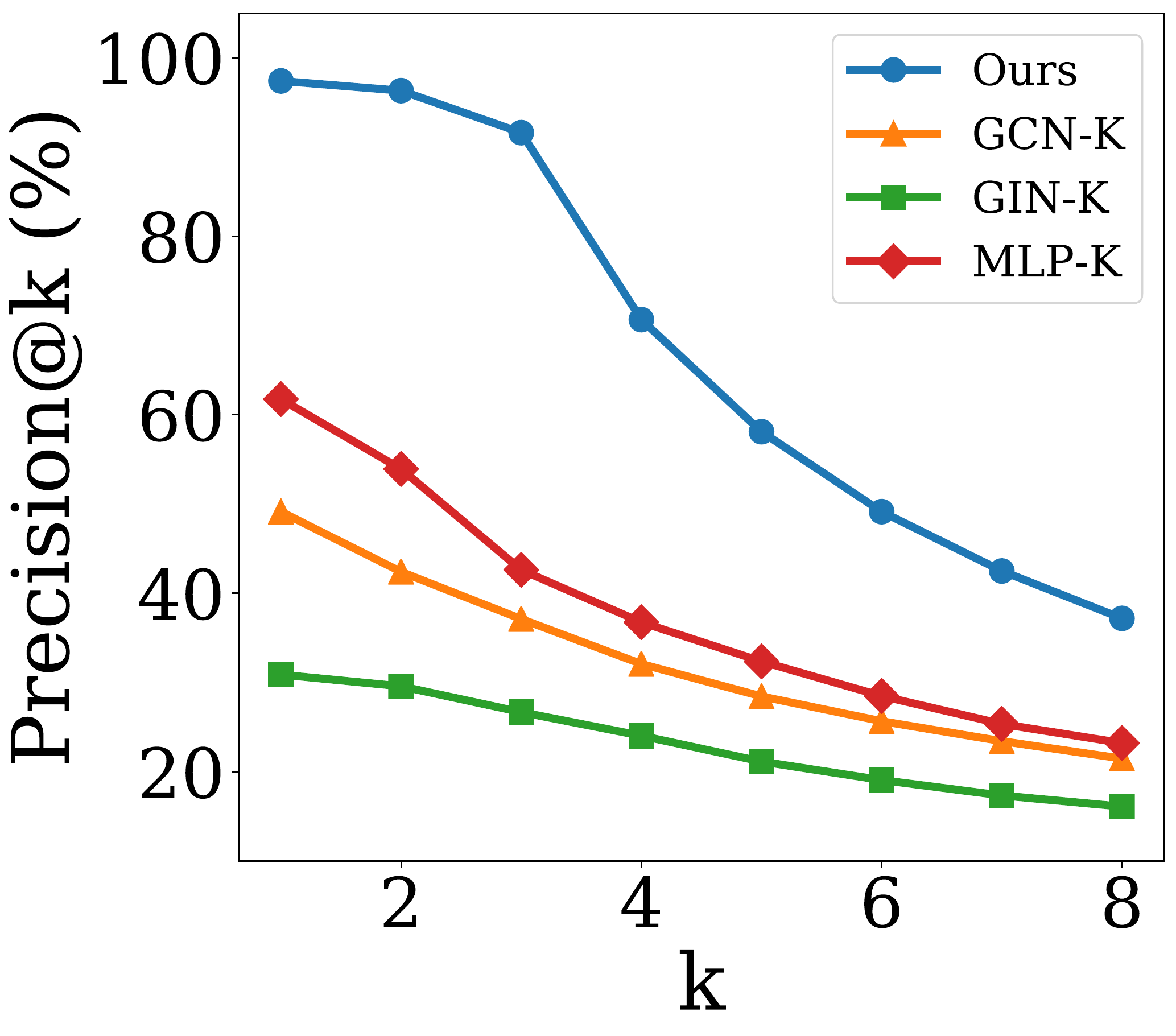} 
    \vskip -0.8em
    \caption{Syn-Cora}
    \label{fig:pre_syn}
\end{subfigure}
\vskip -1.2em
\caption{The precision@k of the $K$-nearest labeled nodes. }
\vskip -1.6em
\label{fig:precision}
\end{figure}

\subsection{Classification and Explanation Quality}
To answer \textbf{RQ1}, we compare {\method} with baselines on real-world datasets and synthetic datasets in terms of classification performance and explanation quality.

\subsubsection{Results on Real-World Datasets} To demonstrate that our {\method} can give accurate predictions, we compare with the state-of-the-art GNNs on real-world datasets. Each experiment is conducted 5 times. The average node classification accuracy and standard deviations are reported in Table~\ref{tab:real_cls}. From the table, we observe:
\begin{itemize}[leftmargin=*]
    \item Our method outperforms GCN and GIN on various real-world datasets especially on the large dataset. This is because information from numerous unlabeled nodes is leveraged in {\method} through the similarity modeling with contrastive learning. 
    \item {\method} achieves comparable performance with self-supervised method SuperGAT. Note that encoder in {\method} only involves 1-hop neighbors to learn representations. This indicates that the designed local structure similarity manages to capture the complex structure information for node classification.
    \item Though GCN-K and GIN-K also utilize GNN to learn representation and $K$-nearest neighbors for prediction, {\method} performs much better than them, which shows the effectiveness of {\method} in utilizing the label information and contrastive learning for learning better representations and similarity metrics.
\end{itemize}

Intuitively, for each test node $v_t \in \mathcal{V}_U$, {\method} should assign higher similarity score to labeled nodes of the same class as $v_t$. To analyze this, for each $v_t$, we first rank the labeled nodes based on similarity scores. Then We treat the label of $v_t$ as the groundtruth and calculate the precision@k for the ranked list. We average the results for all $v_t \in \mathcal{V}_U$. Generally, if a method assigns higher similarity scores to labeled nodes of the same class as $v_t$, it would have large precision@k. We vary $k$ as $\{1,2,\dots,8\}$. The hyperparameters are set as described in Section~\ref{Sec:5.2.2_imp}.
The results on the three real-world datasets are presented in Figure~\ref{fig:pre_cora},~\ref{fig:pre_citeseer} and ~\ref{fig:pre_pubmed}, respectively. We can observe that {\method} consistently outperforms other baselines by a large margin, which indicates that {\method} can retrieve reliable $K$-nearest labeled nodes for prediction and explanation.

\begin{table}[t]
    \small
    \centering
    \caption{Average ratings of human evaluation.}
    \vskip -1.5em
    \begin{tabularx}{0.9\columnwidth}{p{0.12\linewidth}CCCCC}
    \toprule
    Dataset  & MLP-K & GCN-K & GIN-K & Ours\\
    \midrule
    Cora & 0.030 & 0.405 & 0.207 & \textbf{0.763}\\
    Citeseer & 0.030 & 0.311 & 0.326 & \textbf{0.733}  \\
    Pubmed & 0.089 & 0.348 & 0.252 & \textbf{0.674}\\
    \bottomrule
    \end{tabularx}
    \label{tab:real_ex}
    \vskip -1em
\end{table}

\begin{figure}
    \centering
    \includegraphics[width=0.80\linewidth]{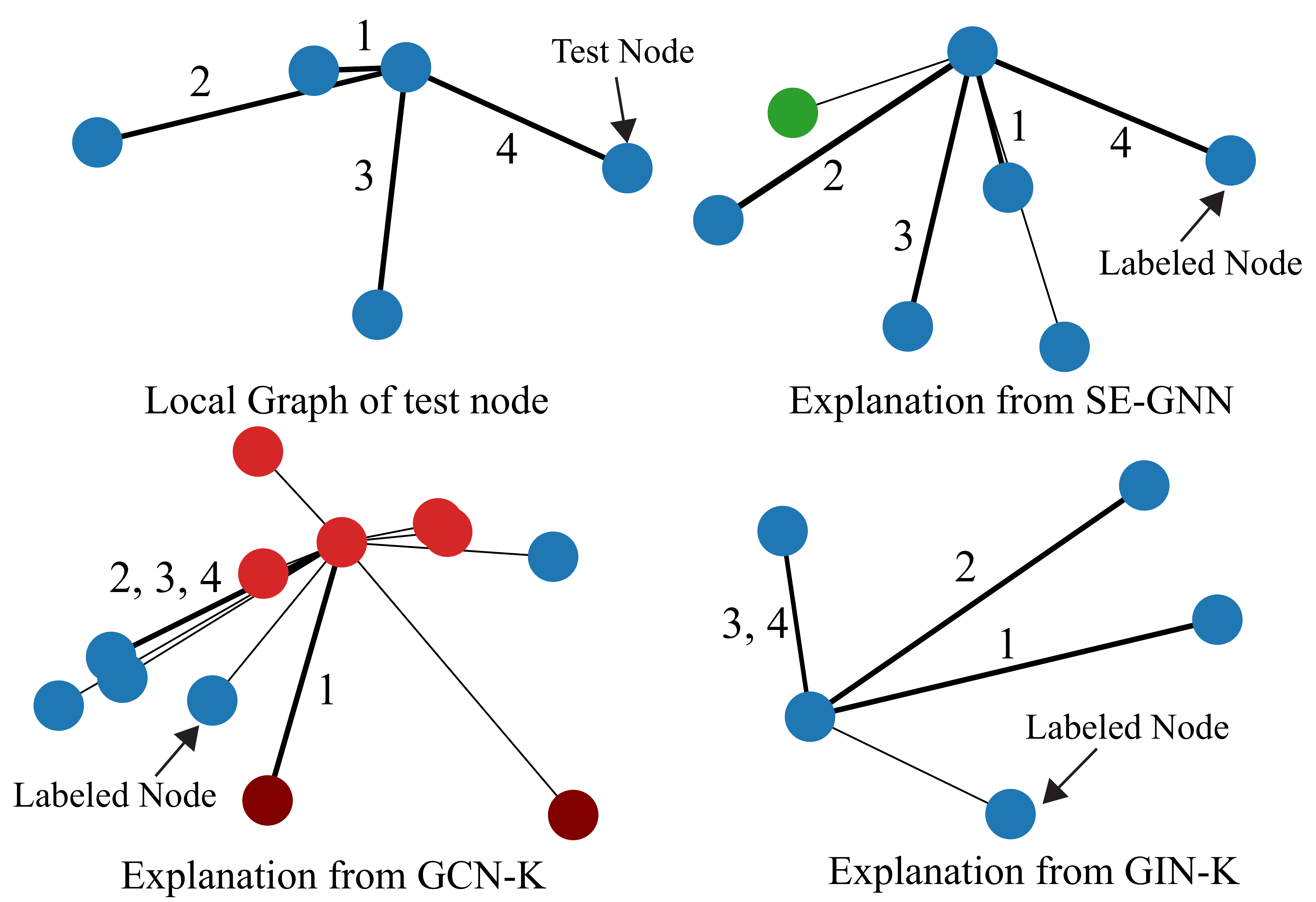}
    \vskip -1em 
    \caption{Illustration of the explanation from {\method} and other baselines. Node colors denote the label of nodes. Edges with the same number denote that they are matched.}
    \vskip -1em
    \label{fig:case}
\end{figure}

We also conduct qualitative evaluation on explanations on real-world datasets. The explanations of an instance from Citeseer are presented in Figure~\ref{fig:case}. Specifically, the local graphs of nearest labeled nodes identified by different methods are presented.
And we apply t-SNE to node features to obtain the positions of nodes in the visualized graph for node similarity comparison. The shapes of the graphs can help to assess the similarity of local structures. From the Figure~\ref{fig:case}, we can observe that SE-GNN can correctly identify the labeled node whose features and local topology are both similar with the target node. And the given edge matching results well explain the local structure similarity. On the other hand, baselines fail to identify the similar labeled nodes and provide poor explanations in structure similarity. 

To further testify the quality of our explanations, 30 annotators are asked to rate the model's explanations on three real-world datasets. The explanations are presented in the same way as Fig.~\ref{fig:case}. Each annotator rates explanations at least 15 instances from the three real-world datasets. The rating score is either 0 (similar) or 1 (disimialr).
The average ratings are presented in Table\ref{tab:real_ex}. From the table, we can find that the nearest neighbors identified by our method receive the highest ratings, which shows that our explanations are in line with the human's decisions. It verifies the quality of our explanations on real-world datasets.
\begin{table}[t]
    \small
    \centering
    \caption{Results on Syn-Cora.}
    \vskip -1.5em
    \begin{tabularx}{0.9\columnwidth}{p{0.22\linewidth}CCCCC}
    \toprule
    Metric (\%)  & MLP-K & GCN-K & GIN-K & Ours\\
    \midrule
    Accuracy & 93.8$\pm 2.3$ & 94.8$\pm 0.7$ & 94.6$\pm 0.7$ & \textbf{97.7}$\pm \mathbf{1.6}$\\
    Edge ACC & - & 25.1$\pm 0.4$ & 18.2$\pm 1.8$ & \textbf{81.1}$\pm 1.1$ \\
    \bottomrule
    \end{tabularx}
    
    \label{tab:syn_cora}
    \vskip -1em
\end{table}
\begin{table}[t]
    \centering
    \caption{Structure explanation AUC on BA-Shapes.}
    \vskip -1.5em
    \begin{tabularx}{0.9\columnwidth}{CCC}
    \toprule
     GNNExplainer & PGExplainer & Ours\\
    \midrule
    95.6$\pm 3.7$ & 98.7$\pm 2.1$ &98.1$\pm 0.5$\\ 
    \bottomrule
    \end{tabularx}
    \label{tab:ba_shape}
    \vskip -1.5em
\end{table}

\subsubsection{Results on Syn-Cora}
We compare with baselines on Syn-Cora which provides the ground-truth explanations to quantitatively evaluate the two-level explanations, i.e., $K$-nearest labeled nodes and the edge matching results for similarity explanation. The prediction performance is evaluated by accuracy. Precision@k is used to show the quality of $K$-nearest labeled nodes. The accuracy of matching edges (Edge ACC) is used to demonstrate the quality of local structure similarity explanation. The results are presented in Table~\ref{tab:syn_cora} and Figure~\ref{fig:pre_syn}. Note that edge matching is not applicable for MLP-K, because it cannot capture structure information. From the table and figure, we observe:
\begin{itemize}[leftmargin=*]
    \item Though GCN-K and GIN-K achieve good performance in classification, they fail to identify the true similar nodes and explain the struck similarity. This is due to the over-smoothing issue in deep GNNs, which leads representations poorly persevere similarity information. By contrast, {\method} achieves good performance in all explanation metrics, which shows node similarity and local structure similarity are well modeled in {\method}.
    \item Compared with MLP-K which does not experience over-smoothing issue, {\method} can give more accurate explanations. This is because we apply the supervision from labels and self-supervision to guide the learning of two-level explanations.
\end{itemize}
\subsubsection{Results on BA-Shapes} As it is discussed in Section~\ref{sec:4.2.1_ex}, our {\method} can be extended to extract a crucial subgraph of the test node's local graph to explain the prediction. To demonstrate the effectiveness of extracting crucial structures as explanations, we compare {\method} with state-of-the-art GNN explainers on a commonly used synthetic dataset BA-Shapes. Following~\cite{ying2019gnnexplainer}, crucial structure explanation AUC is used to assess the performance in explanation. The average results of 5 runs are reported in Table~\ref{tab:ba_shape}. From this table, we can observe that, though {\method} is not developed for extracting crucial subgraph for providing explanations, our {\method} achieves comparable explanation performance with state-of-the-art methods. This implies that accurate crucial structure can be derived from the {\method}'s explanations in local structure similarity, which further demonstrates that our {\method} could give high-quality explanations.

\subsection{Robustness}
\begin{figure}[t]
\centering
\begin{subfigure}{0.49\columnwidth}
    \centering
    \includegraphics[width=0.85\linewidth]{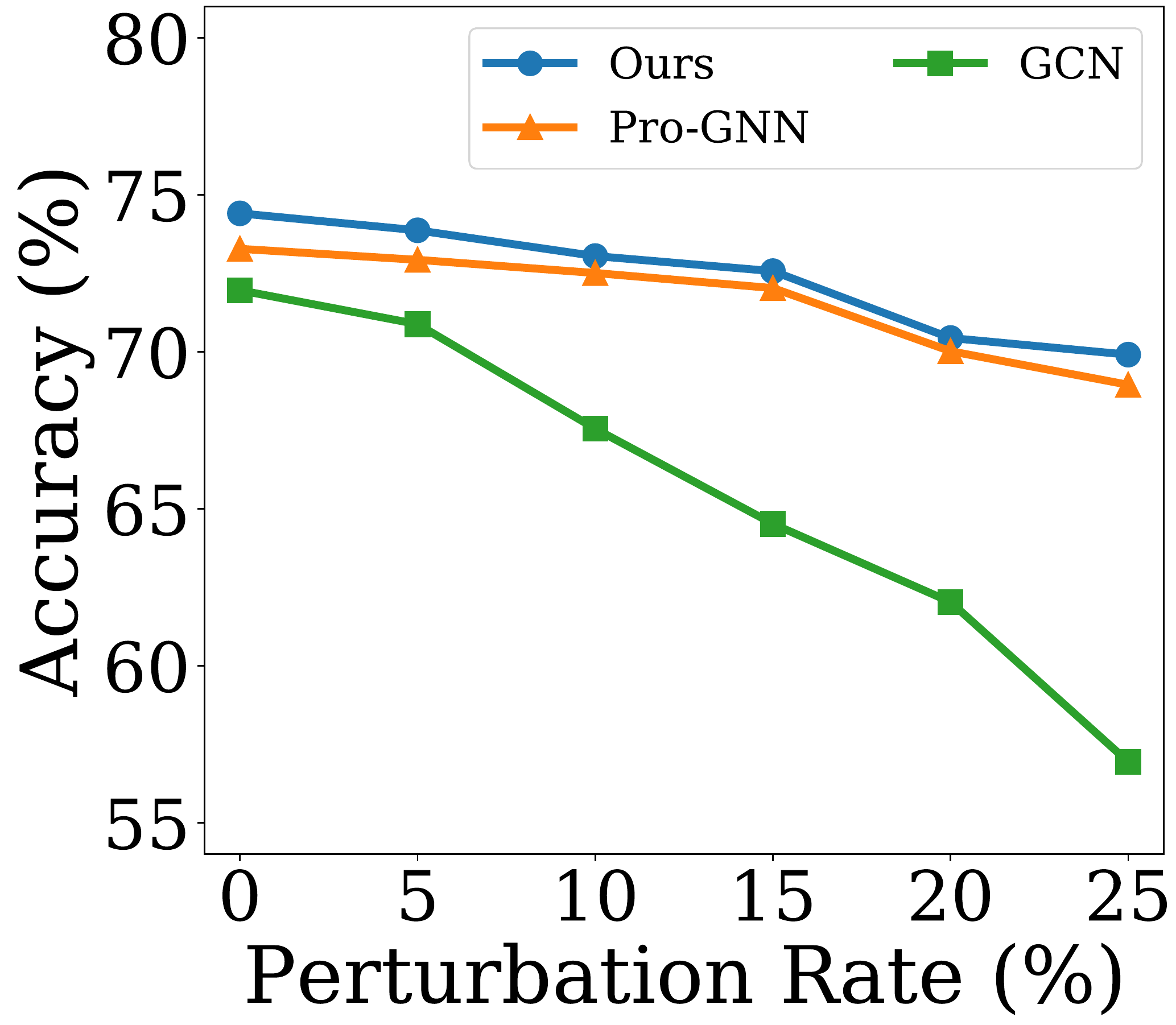} 
    \vskip -0.5em
    \caption{Metattack}
\end{subfigure}~~
\begin{subfigure}{0.49\columnwidth}
    \centering
    \includegraphics[width=0.85\linewidth]{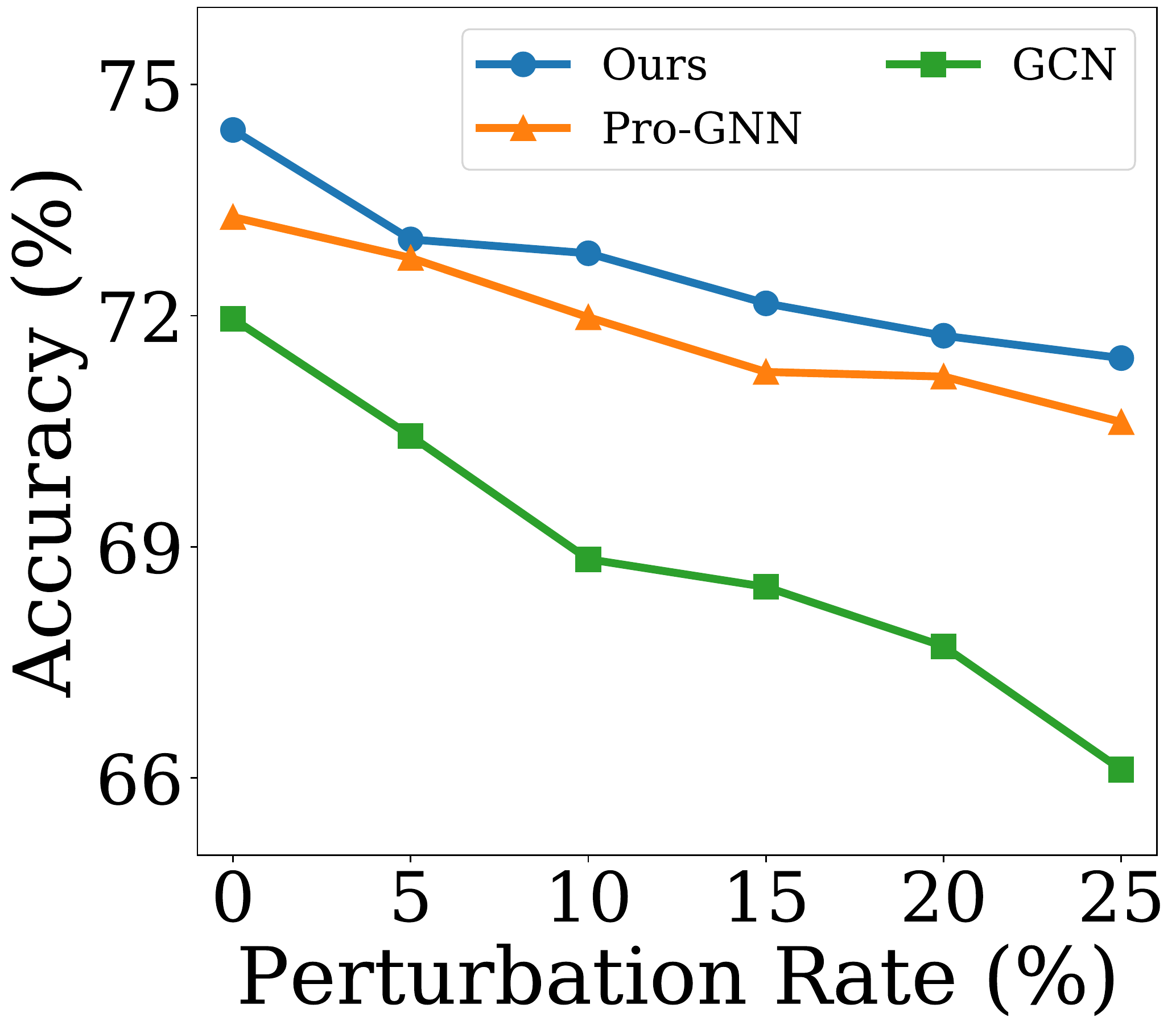} 
    \vskip -0.5em
    \caption{Random Noise}
\end{subfigure}
\vspace{-1.5em}
\caption{Robustness under different Ptb rates on Citeseer. }
\label{fig:ptb}
\vskip -1em
\end{figure}

Structure noises widely exist in the real world and can significantly degrade the performance of GNNs~\cite{zugner2018adversarial,zugner2019adversarial}. {\method} adopts graph topology in representations learning and local similarity evaluation, which could be affected by noisy edges. Therefore, we conduct experiments on noisy graphs to evaluate the robustness of {\method} to answer \textbf{RQ2}.  Experiments are conducted on two types of noisy graphs, i.e., graphs with random noise and non-targeted attack perturbed graphs. For non-targeted attack, we apply \textit{metattack}~\cite{zugner2018adversarial}, which poisons the structure of the graphs via meta-learning. The perturbation rate of non-targeted attack and random noise is varied as $\{0\%, 5\%, \dots, 25\%\}$. The results on Citeseer are shown in Figure~\ref{fig:ptb}. From this figure, we observe that {\method} outperforms GCN by a large margin when the perturbation rates are higher. For example, {\method} achieves over 10\% improvements when the perturbation rate of metattack is 25\%. And {\method} even performs better than Pro-GNN which is one of the state-of-the-art robust GNNs against structure noise. This is because:(i) The contrastive learning in {\method} encourages the representations consistency between the clean graph and randomly perturbed graph. Thus, the learned encoder will not be largely affected by structure noises; (ii) Noisy edges link the nodes that are rarely linked together. Thus, the noise edges generally receive low similarity scores and would not be selected to compute local structure similarity. 

\begin{figure}[t]
\centering
\begin{subfigure}{0.49\columnwidth}
    \centering
    \includegraphics[width=0.85\linewidth]{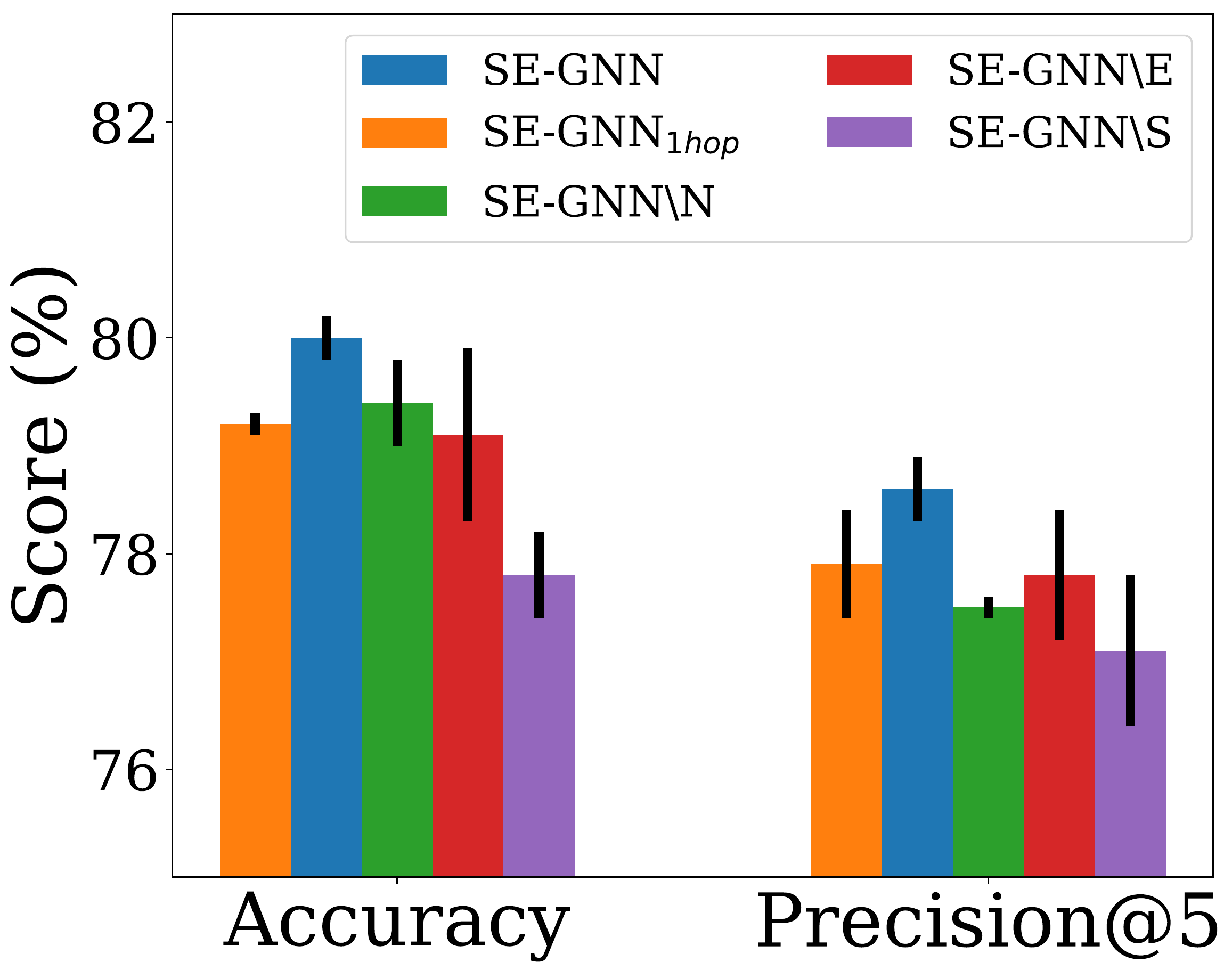} 
    \vskip -0.8em
    \caption{Pubmed}
    \label{fig:abla_pubmed}
\end{subfigure}
\begin{subfigure}{0.49\columnwidth}
    \centering
    \includegraphics[width=0.85\linewidth]{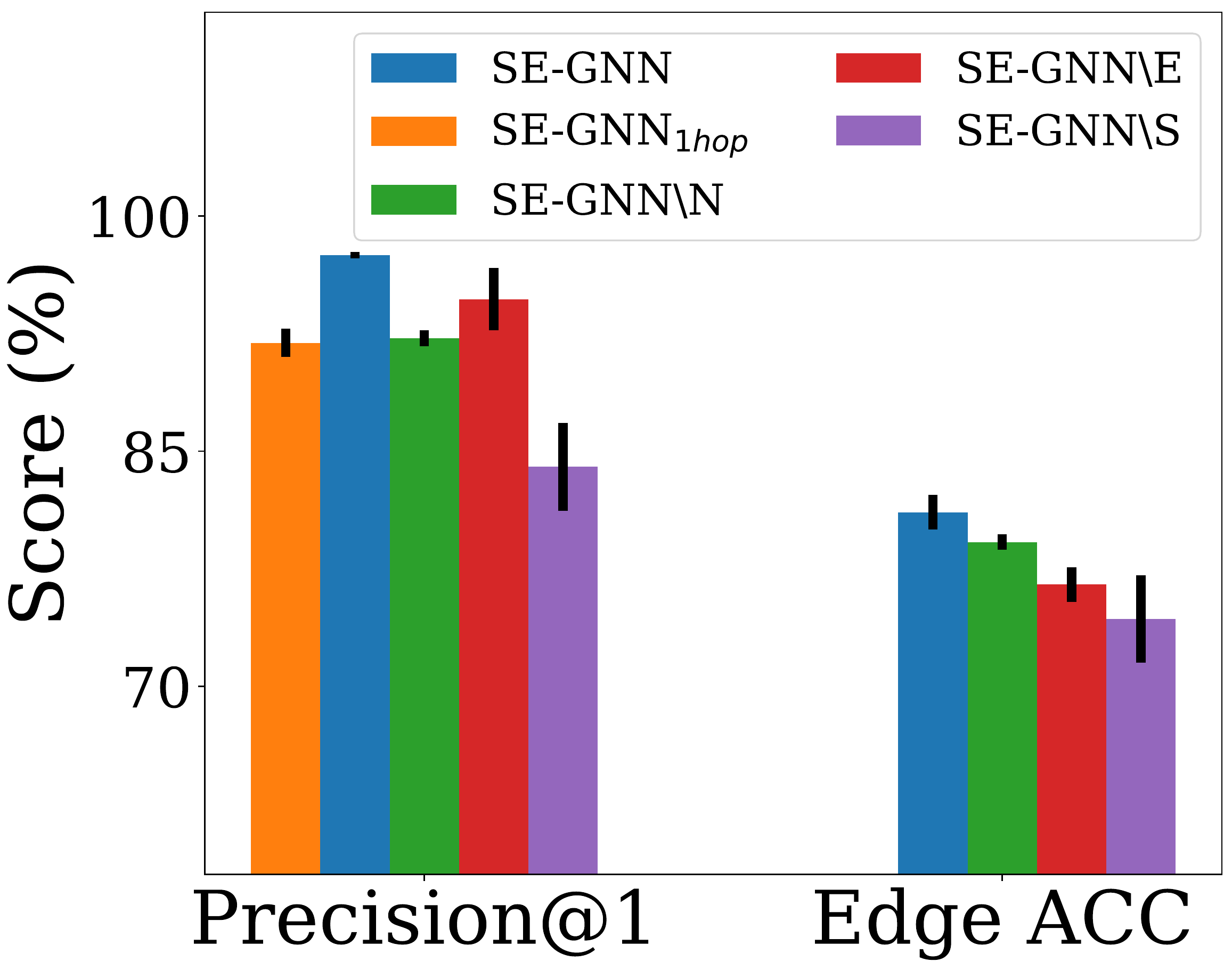} 
    \vskip -0.8em
    \caption{Syn-Cora}
    \label{fig:abla_syn}
\end{subfigure}
\vspace{-1em}
\caption{Comparisons with {\method} and its variants.}
\label{fig:abla}
\vskip -1.5em
\end{figure}

\subsection{Ablation Study}
To answer \textbf{RQ3}, we conduct ablation study to explore the effects of local structure similarity modeling and self-supervision for explanations. {\method} utilizes 2-hop local graphs to obtain local structure similarity. To investigate how the similarity modeling will be influenced by the hop of local graphs, we train a variant {\method}$_{1hop}$ which calculates local structure similarity based on 1-hop local graph. To demonstrate the effectiveness of self-supervision on node similarity, we set $\alpha$ in objective function Eq.(\ref{eq:overall}) as 0 to obtain {\method}$\backslash$N. Similarly, we remove the self-supervision on local structure similarity and obtain a variant named as {\method}$\backslash$E. We also train a variant {\method}$\backslash$S which does not incorporate any self-supervision as the reference. Results on Pubmed and Syn-Cora are presented in Figure~\ref{fig:abla}. Since the edge matching in {\method}$_{1hop}$ only considers 1-hop local graph, Edge ACC on 2-hop local graph is not applicable to {\method}$_{1hop}$.  
From the Figure~\ref{fig:abla}, we can observe that: (i) The performance of {\method}$_{1hop}$ is significantly lower than {\method} in both prediction and explanation. This indicates the importance of incorporating more rich structure information for similarity modeling; and (ii) {\method} outperforms {\method}$\backslash$E and {\method}$\backslash$N by a large margin, which implies that the self-supervision on node similarity and local structure similarity is helpful for identifying interpretable $K$-nearest labeled nodes.
\subsection{Parameter Sensitivity  Analysis}
\begin{figure}[t]
\centering
\begin{subfigure}{0.49\columnwidth}
    \centering
    \includegraphics[width=0.98\linewidth]{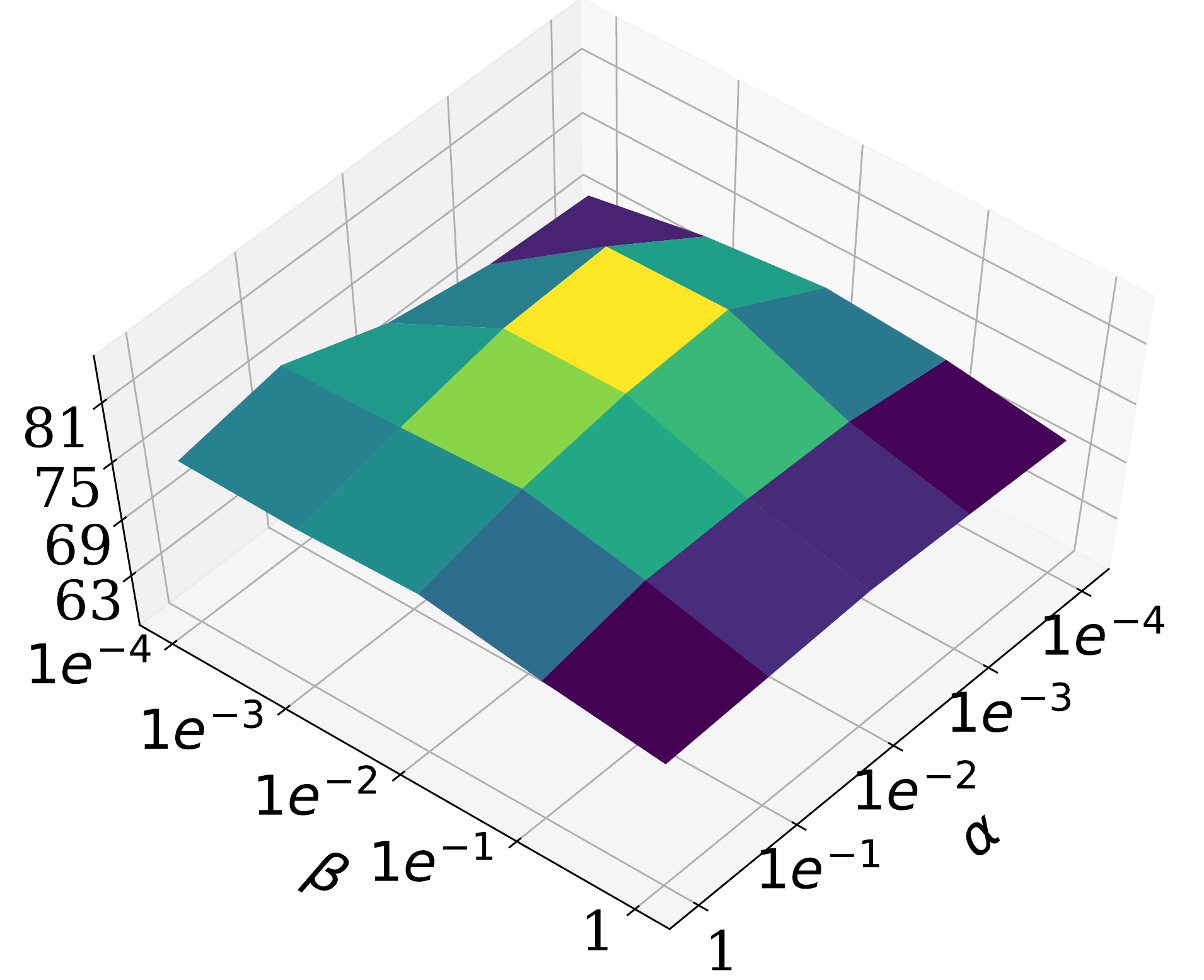} 
    \vskip -0.5em
    \caption{Accuracy (\%)}
\end{subfigure}
\begin{subfigure}{0.49\columnwidth}
    \centering
    \includegraphics[width=0.98\linewidth]{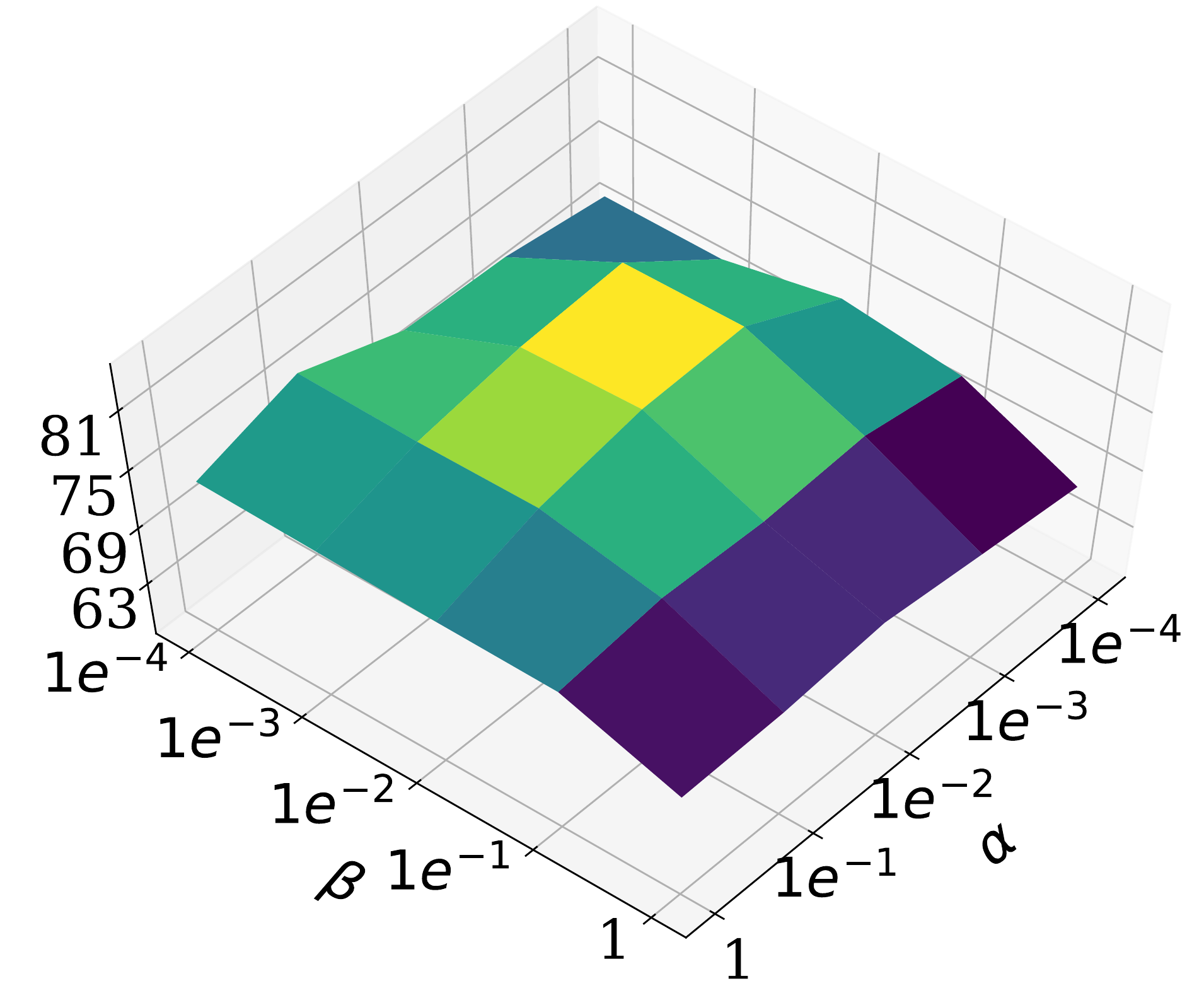} 
    \vskip -0.5em
    \caption{Precision@5 (\%)}
\end{subfigure}
\vspace{-1em}
\caption{Parameter sensitive analysis on Pubmed.}
\label{fig:para}
\vskip -1.5em
\end{figure}

In this subsection, we investigate how the hyperparameter $\alpha$ and $\beta$ affect the performance of {\method}, where $\alpha$ and $\beta$ control the contribution of self-supervision in node similarity modeling and local structure similarity modeling, respectively.  
We vary $\alpha$ and $\beta$ as $\{0.0001, 0.001, 0.01, 0.1 1\}$. We report the classification accuracy and precision@5 of $K$-nearest labeled nodes on Pubmed to show the effects of $\alpha$ and $\beta$ on predictions and explanations, respectively. The results are shown in Fig.~\ref{fig:para}. We find that: (i) with the increase of $\alpha$, the performance in prediction and explanation will first increase and then decrease.  When $\alpha$ is too small, little self-supervision is received for node similarity modeling. Low-quality $K$-nearest labeled nodes are obtained, which results in poor performance in classification and explanation. When $\alpha$ is too large, the overall loss function 
will be dominated by node similarity modeling, which can also lead to a poor overall similarity metric. When $\alpha$ is between 0.001 to 0.01, the performance of {\method} in prediction and explanation is generally good. (ii) Similarly, with the increasing of $\beta$, the performance of {\method} tends to first increase then decrease.  For $\beta$, a value between 0.001 to 0.01 generally gives good performance.

\section{Conclusion and Future Work}
In this paper, we study a novel problem of self-explainable GNNs by exploring $K$-nearest labeled nodes. We propose a new framework, which designs intepretable similarity module for finding $K$-nearest labeled nodes and simultaneously utilizes these nodes for label prediction and explanations. {\method} also adopts the contrastive learning to  benefit the similarity module. Extensive experiments on real-world and synthetic datasets demonstrate the effectiveness of the proposed {\method} for explainable node classification. Ablation study and parameter sensitive analysis are also conducted to understand the contribution of the modules and sensitivity to the hyperparameters. There are several interesting directions which need further investigation. For example, one direction is to extend {\method} for explainable link prediction. Another direction is to investigate other pre-text tasks such as using structural identity~\cite{ribeiro2017struc2vec} as a self-supervision to help the similarity module.

\section{Acknowledgements}
This material is based upon work supported by, or in part by, the National Science Foundation (NSF) under grant  \#IIS1955851, and Army Research Office (ARO) under grant \#W911NF-21-1-0198. The findings and conclusions in this paper do not necessarily reflect the view of the funding agency. 

\bibliographystyle{ACM-Reference-Format}
\bibliography{acmart}

\end{document}